\documentclass{article}

\usepackage{arxiv}

\usepackage[utf8]{inputenc} 
\usepackage[T1]{fontenc}    
\usepackage{hyperref}       
\usepackage{url}            
\usepackage{booktabs}       
\usepackage{amsfonts}       
\usepackage{nicefrac}       
\usepackage{microtype}      
\usepackage{lipsum}
\usepackage{amsmath}
\usepackage{graphicx}
\graphicspath{ {./images/} }

\usepackage{algorithm}
\usepackage{algpseudocode}
\usepackage{amsmath}
\usepackage{graphicx} 
\usepackage{amsmath}
\usepackage{amsfonts}
\usepackage{multirow}
\usepackage{multicol}
\usepackage[table]{xcolor}
\usepackage{booktabs}
\usepackage{multirow}
\usepackage{siunitx}
\usepackage{array}
\usepackage{booktabs, multirow, array}
\usepackage{makecell} 

\title{Important Equations}
\author{itxwaleedrazzaq }
\date{March 2025}

\usepackage{amsmath,amsfonts}
\usepackage{algpseudocode}
\usepackage{algorithm}
\usepackage{array}
\usepackage[caption=false,font=normalsize,labelfont=sf,textfont=sf]{subfig}
\usepackage{textcomp}
\usepackage{stfloats}
\usepackage{url}
\usepackage{verbatim}   
\usepackage{multirow}
\usepackage{rotating}
\usepackage{graphicx}
\usepackage{multicol}
\usepackage{multirow}
\usepackage{booktabs}    
\usepackage{cellspace}
\usepackage{subcaption}

\title{A Novel Multimodal RUL Framework for Remaining Useful Life Estimation with Layer-wise Explanations}

\author{
  Waleed Razzaq \\
  School of Automation\\
  University of Science and Technology China\\
  Hefei, Anhui \\
  \texttt{waleedrazzaq@mail.ustc.edu.cn} \\
   \And
 Yun-Bo Zhao \thanks{Corresponding author. Email: \texttt{ybzhao@ustc.edu.cn}} \\
  School of Automation\\
  University of Science and Technology China\\
  Hefei, Anhui \\
  \texttt{ybzhao@ustc.edu.cn} \\
}

\begin{document}
\maketitle

\begin{abstract}
 \textbf{Background:} Estimating Remaining Useful Life (RUL) of mechanical systems is a critical aspect of Prognostics and Health Management (PHM) systems. Accurate RUL prediction enables timely maintenance, enhances system reliability, and reduces operational costs. Among various mechanical components, rolling-element bearings are a leading cause of industrial machinery failure, underscoring the need for robust and reliable RUL estimation techniques.\\
 \textbf{Problem:} Despite numerous existing approaches, many suffer from limitations such as poor generalizability, lack of robustness, high data requirements, and limited interpretability. \\
\textbf{Methods:} To address these challenges, this paper proposes a novel multimodal-RUL framework that integrates both image representations (ImR) and time-frequency representations (TFR) of one-dimensional nonstationary vibration signals obtained from multichannel accelerometers. The proposed multimodal-AI architecture comprises three distinct branches. The \textit{image} and \textit{TFR branches} extract spatial degradation features using multiple dilated convolutional blocks with residual connections, applied to the ImR and TFR inputs, respectively. These features are then merged in \textit{fusion branch}, where a Long Short-Term Memory (LSTM) network captures temporal degradation patterns. A multi-head attention (MHA) mechanism further refines these features by emphasizing the most informative aspects, which are subsequently processed through Linear layers for RUL prediction. To facilitate effective multimodal-learning, we introduce a comprehensive feature engineering framework. Vibration signals are rasterized into ImR using the Bresenham Line Drawing algorithm, while TFRs are derived via Continuous Wavelet Transform. Additionally, we propose a layer-wise relevance propagation method tailored for multimodal architectures (\textit{multimodal-LRP}), enhancing model transparency and interpretability. \\
\textbf{Results:} The effectiveness of the proposed method is validated using the XJTU-SY and PRONOSTIA benchmark datasets. Experimental results demonstrate that our approach matches and sometimes outperforms baseline models under both seen and unseen operating conditions. Notably, the model achieves competitive performance while requiring approximately 28\% less training data for XJTU-SY and 48\% less for PRONOSTIA. The results also suggest that the proposed model exhibits strong resilience to various types of noise. Furthermore, the multimodal-LRP explanations substantiate the interpretability and trustworthiness of the method, reinforcing its practical applicability in industrial settings.

\end{abstract}


\section{Introduction}
Prognostic Health Management Systems (PHMs) are an integral part of complex industrial systems, ensuring safety and reliability by continuously monitoring and evaluating the health conditions of critical components. PHMs are essential for preventing severe operational hazards and ensuring accident-free processes. Remaining Useful Life (RUL) is a key function of PHMs that determines the residual operational lifespan of a machine or its components. Nonstatic machines, particularly rotating machines with rolling-element bearings, are more susceptible to failure due to operation under extreme conditions, leading to wear and tear that affects performance. Several studies have indicated that approximately 40--50$\%$ of industrial machine failures are related to these bearings \cite{ding2021remaining}. Therefore, an accurate RUL estimation system is essential for monitoring effective conditions, mitigating risk, and preventing unexpected breakdowns that could disrupt production. In recent years, there has been a focus on developing RUL systems for rolling-element bearings, and several notable accomplishments have been observed, broadly categorized into two approaches: physics-based and data-driven.

Physics-based modeling provides information on bearing degradation processes via a set of equations derived from mathematical representations of physical systems. Gazizulin et al. \cite{gazizulin2018physics} proposed a hybrid approach that integrates finite element (FE) modeling with damage mechanics and experimental validation utilizing continuum damage mechanics to simulate spall formation on bearing raceways, incorporating microstructural details through Poisson‒Voronoi tessellation. To enhance accuracy, nonlinear dynamic modeling with FE simulations is combined to analyze the stress‒strain response at the spall edge where the rolling elements repeatedly impact the surface. Gabrielli et al. \cite{gabrielli2024physics} introduced a degradation model based on equivalent damaged volume (EDV) that combines experimental run-to-failure data with a lumped parameter model to estimate defect growth. The EDV algorithm quantifies the equivalent damaged volume of the defect by comparing experimentally acquired vibration signals with numerically simulated signals, taking into account the defect depth and angular extent. Guo et al. \cite{guo2023dynamics} proposed a dynamic model for a rolling bearing variable stiffness system with local faults, focusing on the support provided by load-carrying rolling elements at fault positions. The model accounts for contact deformation, retention factors, and analyzes changes in effective contact stiffness, deformation, contact force, and total effective stiffness. Ni et al. \cite{ni2024dynamic} developed a finite element dynamic analysis model for rolling bearings with outer ring pit defects, considering factors such as load, speed, contact, and friction. The model analyzes the dynamic response of the bearing, focusing on vibration characteristics caused by the defect. It provides insights into how outer ring defects influence bearing degradation under varying operating conditions. Although successes have been observed, there are still challenges associated with this approach. The development of these models requires extensive interdisciplinary expertise and complex mathematical formulation. Even with such expertise, capturing the exact system’s behavior remains a challenge. Another limitation is their inability to account for real-world disturbances unless explicitly modeled within the equations, making real-world applications more limited than desired.

Data-driven methods identify hidden patterns in condition monitoring data, enabling a more accurate assessment of system performance and potential failures. Cheng et al. \cite{cheng2020deep} proposed a framework that combined a CNN with Hilbert-Huang Transform (HHT) and Epsilon-Support Vector Regression ($\varepsilon$-SVR). This extracts a nonlinear Degradation Energy Indicator (DEI) from vibrational signals to learn the relationship between the DEI and raw vibration signals through CNN, resulting in enhanced prediction accuracy. Finally, $\varepsilon$-SVR is employed to predict the RUL based on the estimated DEI of the test rolling-element bearings. While effective, this approach faces limitations such as dependence on preferred degradation indicators, the assumption of constant operating conditions, and difficulties in generalizing to unseen fault patterns. Luo et al. \cite{luo2022convolutional} proposed a Convolutional-Based Attention Mechanism BiLSTM (CABiLSTM) that integrates a CNN for feature extraction and Bidirectional-LSTM with an attention mechanism to better capture long-term time‒frequency relationships. Instead of using the standard multiplication operation, CABiLSTM applies convolutional operations directly to the Bi-LSTM cell states, improving the representation of features. However, this approach has limitations such as difficulties in detecting early failures, a strong dependence on CNN feature extraction, and the need for further fine-tuning to enhance model generalization. Yang et al. \cite{yang2022novel} proposed a multistep approach that integrates signal decomposition, feature extraction, and neural network-based forecasting. Initially, vibration signals are decomposed via piece-wise cubic Hermite interpolating polynomial–local characteristics-scale decomposition (PCHIP-LCD), with effective intrinsic scale components (ISCs) selected via the kurtosis‒correlation coefficient (K‒C) criterion. These ISCs are further enhanced through an improved independent component analysis (IICA) and the Mahalanobis distance (MS), resulting in a sensitive degradation indicator (IICAMD). The gray regression neural network (GRNN) is used to smooth out fluctuations in IICAMD for the final RUL estimation. This approach is limited by potential overfitting and difficulty in handling varying operating conditions. Zhang et al. \cite{zhang2023integrated} introduced an integrated multi-head dual-parse self-attention network (IMDSSN) leveraging a modified transformer architecture for RUL estimation. To improve the efficiency of traditional transformers, the IMDSSN employs a multi-head probability sparse self-attention network (MPSN) that eliminates unnecessary dot-product operations, reducing computational overhead. Additionally, a multi-head logSparse self-attention (MLSN) is introduced to calculate attention weights logarithmically, resulting in enhanced RUL estimation. However, the model faces challenges in generalizability and requires highly refined training data with a consistent distribution, limiting its applicability for cross-domain predictions.

Zhao et al. \cite{zhao2023multi} proposed the MSIDSN framework designed to process multisensory degraded data across various scales by integrating a multiscale CNN block with a self-attention mechanism, a recurrent module, and a feature extraction technique. This approach extracts multisensory-temporal features by leveraging interdependencies and facilitating mutual interaction. While it enhances prediction accuracy through efficient loss, its performance is affected by variations in sensor quality and the presence of noise in the data. Gao et al. \cite{gao2024long} proposed an adaptive stage division neural network using augmented continuous wavelet transform to identify degradation starting points in rolling bearings. A temporal attention-based neural network (LTAN) with multilevel dilated causal convolutions was designed to capture long-term degradation patterns. The complete AD-LTAN framework integrates preprocessing with EMD, adaptive stage detection, and temporal attention mechanisms for accurate RUL prediction under variable conditions.
However, the method's increased complexity due to dual temporal modules may limit real-time deployment in resource-constrained environments. Wen et al. \cite{wen2024early} proposed a two-step method for early prediction of rolling bearing RUL by integrating an envelope spectral indicator (ESI) with an extended Kalman filter (EKF). ESI was constructed from fault characteristic frequencies in the averaged envelope spectrum to isolate bearing-specific degradation signals. EKF was then applied to the ESI series to recursively estimate the nonlinear exponential degradation process and predict RUL. The method's accuracy in early-stage prediction may degrade due to sensitivity to initial EKF parameters and non-monotonic ESI behavior. Magdan et al. \cite{magadan2024robust} proposed a robust health prognostics methodology combining time- and frequency-domain vibration feature extraction, a stacked variational denoising autoencoder (SVDAE) for feature fusion and Health Indicator construction, and a bidirectional LSTM (BiLSTM) network for RUL prediction, validated across diverse motor conditions without retraining or fine-tuning. Despite their limitations, data-driven methods face several common challenges that must be addressed. One major issue is that the generalization ability and reliability of these models heavily depend on the quality of the extracted features. However, many existing frameworks lack robust and compact feature engineering pipelines, leading to inconsistencies and scalability issues. Second, these models require massive amounts of refined and diverse data to perform well and generalize effectively. Third, the performance of such systems typically degrades as input perturbations increase—a significant problem in real-world scenarios where environmental noise can affect the quality of sensor data. Furthermore, these approaches often operate as black-box models, providing predictions without offering insight into how or why a particular outcome was reached. This lack of interpretability makes validation difficult and reduces trust in the system, especially in high-risk industrial applications where accountability is critical. The inability to explain model behavior remains a significant concern, highlighting the need for more interpretable and explainable AI-driven prognostic solutions.

To overcome the identified challenges, we propose a novel multimodal-RUL framework that leverages ImR and TFR of multichannel nonstationary vibrational data acquired from the accelerometer. Our framework consists of three distinct branches: \textit{image branch}, \textit{TF branch}, \textit{fusion branch}. The image and TF branches operate in parallel to process spatial degradation patterns via multiple convolutional blocks, each comprising dilated convolutional layers designed to extract spatial degradation patterns from ImR and TFR, respectively. The first block employs CNN layers with small filter sizes and high dilation rates, capturing intricate local features while preserving large receptive fields and enabling the model to detect subtle degradation patterns. In contrast, the second block uses larger filter sizes with low dilation rates to capture large-scale degradation patterns and global spatial correlation, which are crucial for accurate RUL estimation. The fusion branch combines the outputs from the image and TF branches through concatenation and feeds them into the LSTM block to capture long-term temporal degradation patterns. To enhance reliability and reduce computational complexity, we introduced residual connections between the CNN and LSTM blocks, ensuring smooth gradient flow during training. The extracted temporal degradation patterns are then processed by a multi-head attention mechanism (MHA), which improves the model's ability to focus on the most relevant patterns while suppressing redundant information crucial for filtering perturbations. The extracted features are then passed through a series of linear layers for final RUL prediction. We also propose a comprehensive and robust multimodal feature engineering framework that converts multichannel nonstationary vibration signals into 2D-ImR via the Bresenham line drawing algorithm and TFR via the wavelet transform. To ensure the interpretability and trustworthiness of the proposed framework, we also propose an enhanced multimodal layerwise relevance propagation (multimodal-LRP) capable of handling advanced multimodal neural networks to provide a localized explanation of model predictions. The highlights of this research are listed below:
\begin{itemize}
    \item A multimodal feature engineering framework is proposed to convert 1D multichannel nonstationary signals into ImR via the Bresenham line drawing algorithm and TFR via the wavelet transform.
    \item A first-of-its-kind multimodal RUL framework is proposed to estimate the RUL of rolling-element bearings. The multimodal AI architecture comprises three branches: \textit{image branch} that extracts spatial degradation patterns from ImR, \textit{TF branch} that captures spatial degradation patterns from TFR, \textit{fusion branch} that integrates these patterns and feeds them into \textit{fusion branch} for modeling long-term dependencies through LSTM and filtering perturbations through MHA and estimating RUL through Linear layers.
    \item To ensure transparency and trust in our model, we propose a multimodal layerwise relevance propagation (multimodal-LRP) algorithm capable of handling advanced multimodal neural networks with both ImR and TFR inputs simultaneously to provide relevance.
    \item We validate the effectiveness of our proposed frameworks using the XJTU-SY and PRONOSTIA bearing datasets, demonstrating their robustness and applicability in real-world industrial settings.
\end{itemize}

The remainder of the paper is organized as follows: Section \ref{sec:prelim} outlines the preliminaries and building blocks of our proposed framework. Section \ref{sec:method} presents the detailed methodology, including the comprehensive feature engineering framework, multimodal-AI and multimodal-LRP algorithm. Section \ref{sec:data_exp} describes the dataset utilized and the training optimization techniques applied to enhance the proposed AI framework. Section \ref{sec:results} discusses the experimental results, validating the network against itself, baseline models, and providing localized multimodal-LRP explanations. Section \ref{sec:discussion} discuss the limitations and possible future research enhancement of our work. Section \ref{sec:conclude} concludes the paper by addressing key findings.

\section{Preliminaries}\label{sec:prelim}
\subsection{Bresenham's line algorithm}
\begin{figure*}[ht!]
\centering
\includegraphics[width=0.7\textwidth]{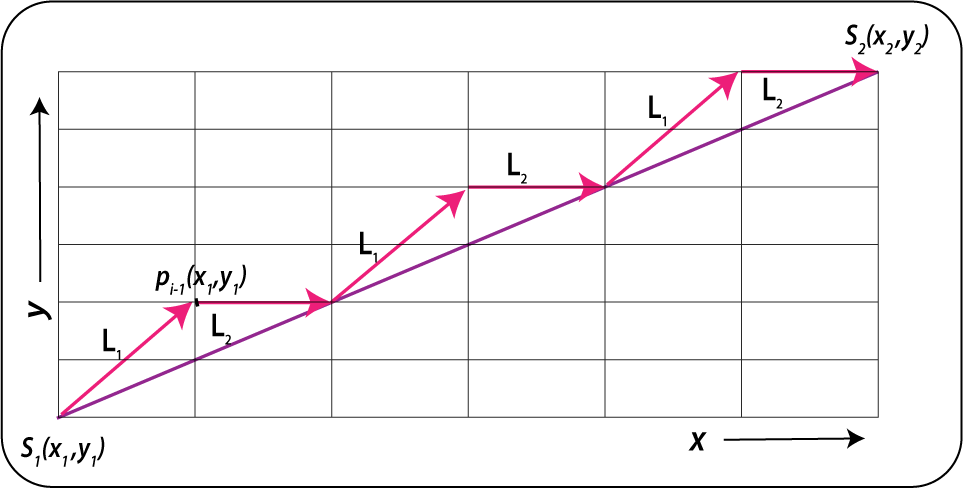}
\caption{Illustration of Bresenham's line algorithm.}
\label{fig:line}
\end{figure*}

Bresenham’s line algorithm is an efficient rasterization technique for approximating discrete linear paths between two points, $(S_1(x_1,y_1))$ and $(S_2(x_2,y_2))$, using integer coordinate selection (see Figure \ref{fig:line}). By iteratively determining the nearest raster position, it achieves accurate linear interpolation without floating-point arithmetic, significantly reducing computational overhead compared with alternative approaches. The algorithm works by first computing the coordinate differences $\Delta x = x_2 - x_1$ and $\Delta y = y_2 - y_1$ and initializing a decision parameter $\rho_0$. Two types of decision parameters can be expected: shallow slopes ($|\Delta y| \leq |\Delta x|$) and steep slopes ($|\Delta y| > |\Delta x|$). For shallow slopes, it initializes the decision parameter $\rho_0 = 2\Delta y - \Delta x$, where at each step, the parameter $\rho$ is evaluated. If $\rho < 0$, the next point is selected along the primary axis $L_2$; if $\rho > 0$, the point advances diagonally to $L_1$, updating as $\rho_0 = \rho + 2\Delta y - \Delta x$. For steep slopes (where axes are swapped to iterate along the vertical axis), the initial decision parameter is $\rho_0 = 2\Delta x - \Delta y$ with analogous updates. This method ensures a continuous, optimal linear approximation by incrementally adjusting decision parameters with minimal error accumulation \cite{bresenham1998algorithm}.

\subsection{Continuous Wavelet Transform}

The continuous wavelet transform (CWT) is a powerful mathematical tool used to decompose a time-varying signal into highly localized oscillations called wavelets, providing better time-frequency analysis. The CWT uses basis functions that are scaled and shifted versions of the time-localized wavelet, enabling the creation of a time-frequency representation of a signal with excellent localization in both time and frequency. The mathematical expression of the CWT is as follows:
\begin{align}
\Gamma(a,b) = \int_{-\infty}^{\infty} I(t) \psi^* \left(\frac{t - b}{a} \right) dt
\end{align}

where \(\Gamma(a,b) \) represents the wavelet coefficients at scale \(a \) and translation \(b \), \(I(t) \) represents the nonstationary signal, and \(\psi(t) \) represents the mother wavelet function. We opted for the Morlet wavelet \cite{ngui2013wavelet} because of its favorable trade-off between time and frequency resolutions, with the frequency resolution improving at higher values of \(a \), and vice versa \cite{lin2000feature}. The Morlet wavelet is defined as a sinusoidal function modulated by a Gaussian envelope with a central frequency \(f_c \) and is given by:
\begin{align}
\psi(t) = e^{\frac{if_c t}{2\pi}} e^{-t^2/2}
\end{align}

\subsection{Layerwise Relevance Propagation (LRP)}

Layerwise relevance propagation (LRP) is a local explainability technique designed to interpret complex deep neural networks by redistributing the output prediction backward through neural networks layer by layer \cite{montavon2019layer}. The process assigns a relevance score to each input feature, indicating its contribution to the final prediction. The LRP operates on the principle of relevance conservation, meaning that the total relevance remains preserved across layers. If $R_j$ represents the relevance of a neuron $j$ in a layer and $R_i$ is the relevance of neuron $i$ in the previous layer, the following relationship holds:
\begin{equation}
    \sum_i R_i^{(l)} = \sum_j R_j^{(l+1)}
\end{equation}
Given a neural network with layers $l = 1,2,\dots,L$, let $x_i^{(l)}$ denote the activation of neuron $i$ in layer $l$ and let $w_{ij}^{(l)}$ represent the weight connection of neuron $i$ in layer $l$ to neuron $j$ in layer $l+1$. The relevance $R_j^{(l+1)}$ is distributed across layers based on specific propagation rules. Here, we focus on three fundamental rules:

\textbf{LRP-0}: This rule is used for basic relevance propagation in cases where the activations and weights are non-negative:
\begin{equation}
    R_i^{(l)} = \sum_j \frac{x_i^{(l)} w_{ij}^{(l)}}{\sum_k x_k^{(l)} w_{kj}^{(l)}} R_j^{(l+1)}
\end{equation}

\textbf{LRP-$\epsilon$}: This rule adds a small stabilizer $\epsilon$ to avoid division by zero in the basic rule:
\begin{equation}
     R_i^{(l)} = \sum_j \frac{x_i^{(l)} w_{ij}^{(l)}}{\sum_k x_k^{(l)} w_{kj}^{(l)} + \epsilon \, \cdot \text{sign} \left(\sum_k x_k^{(l)} w_{kj}^{(l)} + \epsilon \right)} R_j^{(l+1)}
\end{equation}

\textbf{LRP-$\gamma$}: This rule separates positive and negative contributions:
\begin{equation}
     R_i^{(l)} = \sum_j \frac{x_i^{(l)} \left(w_{ij}^{(l)} + \gamma w_{ij}^{(l)+} \right)}{\sum_k x_k^{(l)} w_{kj}^{(l)} + \gamma w_{ij}^{(l)+}} R_{j}^{(l+1)}
\end{equation}

\section{Methods}\label{sec:method}

\subsection{Multimodal Feature Engineering Framework}
The condition monitoring signal of rolling-element bearings typically comprises one-dimensional (1-D) nonstationary signals from a multichannel accelerometer. It is essential to establish a comprehensive feature engineering framework to extract discriminative features for training AI systems. Assuming that the number of channels is \(N_c \) and the data length is \(L_d \), the raw data can be represented as
\begin{equation}
I(t) = [i_1, i_2, \dots, i_{L_d}], \quad i_k = [n_{1k}, n_{2k}, \dots, n_{N_{c}k}]
\end{equation}
where \(i_k \) represents the sensor's reading at time step \(k_k \). To extract useful local features, it is essential to break the signal into smaller segments using the windowing technique as follows:
\begin{equation}
    i_w(t) = I(t) \cdot w_i(t)
\end{equation}
where \(w_i(t) \) is the window function with window length \(L_w \) for the \(i-\)th segment defined as
\begin{equation}
    w_i(t) =
\begin{cases}
1, & \text{if } t \in [t_i, t_{i+1} + L_w] \\
0, & \text{otherwise}
\end{cases}
\label{eq:window}
\end{equation}

\subsubsection{Image Representation (Rasterization)}
Rasterization transforms vector-based representations into discrete pixel arrays, which are essential for visualizing time series signals as 2D images. Before rasterization, the raw 1D signals are normalized to the range $[0,1]$ via min-max normalization:

\begin{equation}
x' = \frac{x - \min(\mathbf{x})}{\max(\mathbf{x}) - \min(\mathbf{x})}
\label{eq:minmax}
\end{equation}

where \(x'\) is the normalized value, and \(\min(\mathbf{x})\) and \(\max(\mathbf{x})\) denote the global minimum and maximum of the dataset, respectively. In our approach, normalized signals are segmented into a window of 1,000 samples (\(L_w = 1000\)). To map these windows into a 2D raster image, the y-axis encodes the signal amplitude at a resolution of $0.015625$ per pixel, whereas the x-axis corresponds to sample indices, resulting in images of size $(64 \times 1000)$. To optimize the computational complexity, we introduce one dummy pixel between adjacent true sample points, effectively approximating the signal’s trajectory while keeping the algorithmic complexity low \cite{patel2023non}. Bresenham’s line algorithm is applied to connect these discrete samples via linear segments, ensuring minimal quantization artifacts and geometric continuity in the resulting 2D image representation (ImR). (See Algorithm \ref{algo:rast}).

\begin{algorithm}
\caption{Rasterization Algorithm}\label{algo:rast}
\footnotesize
\begin{algorithmic}[1]
\Require horizontal channel windowed segments $I_{wh}$, window length $L_w$, y-resolution ($y_{res}$), image dimensions ($image_{height}, image_{width}$), dummy spacing ($dummy$)
\Ensure Rasterized image of size ($image_{height}, image_{width}$)

\State $image \gets \text{empty array of size } (image_{height}, image_{width})$

\For{$i_w$ in $I_{wh}$}
    \State $y1 \gets \text{map values in } window[:-1]$
    \State $y2 \gets \text{map values in } window[1:]$
    \State $x1 \gets \text{arange}(0, L_w \times dummy, dummy)[:-1]$
    \State $x2 \gets \text{arange}(dummy, L_w \times dummy + 1, dummy)[:-1]$
    \State $pairs \gets \text{zip}(x1, y1, x2, y2)$
    \For{each pair $(x_1, y_1, x_2, y_2)$ in $pairs$}
        \State $img_x, img_y\gets \textbf{\textit{Bresenham}}(x_1, y_1, x_2, y_2)$
        \State $image[img_y, img_x] \gets 255$
    \EndFor
\EndFor
\State \Return $image$
\end{algorithmic}
\end{algorithm}

\subsubsection{Time‒Frequency Representations (TFR)}
The window signal \(i_w(t) \) of \(L_w = 1,000\) is passed to the CWT to capture localized time-frequency features vital for accurately modeling degradation patterns for accurate RUL estimation. The CWT can be mathematically represented as
\begin{equation}
    \Gamma_{iw}(a,b) = \int_{-\infty}^{\infty} i_w(t) \psi^* \left(\frac{t - b}{a} \right) dt
    \label{eq:wavelet}
\end{equation}
where \(\Gamma_{iw}(a,b) \) represents the wavelet coefficient of the windowed signal. \(\psi \) is the Morlet wavelet. To extract useful features from the CWT, it is imperative to carefully select the frequency of interest \((f_{\min}, f_{\max})\) to define the scale range \(a\) of the CWT. The choice of the range is usually informed by the system's operating frequency \(f_o \), allowing the model to accommodate multiple scenarios effectively. We considered up to the third harmonic of the operating frequency, providing a good balance between efficiency and capturing useful features ($f_\text{min} \approx f_o/3, \textit{ }f_\text{max} \approx 3f_o)$. Then, the corresponding wavelet transform scales can be calculated as
\begin{equation}
    a_{\min} = \frac{f_c}{f_{\max} \cdot f_{\text{sampling}}}, \quad a_{\max} = \frac{f_c}{f_{min} \cdot f_{\text{sampling}}}
\end{equation}
where \(f_{\text{sampling}} \) is the sampling frequency of the vibrational signal, and \(f_c \) is the central frequency of the Morlet wavelet, typically chosen as \(f_c = 0.81 \), which governs the trade-off between time and frequency resolution. To ensure comprehensive coverage of the frequency range, logarithmically spaced scales are used.
\begin{equation}
    a_i \in [a_{\min}, a_{\max}], \quad i = 1,2, \dots, N
\end{equation}
where \(N \) is the number of scales selected based on the desired resolution in the TF domain. The following TF features are derived (see Algorithm \ref{algo:tf}), which represent the system state \cite{razzaq2025carle}:
\begin{itemize}
    \item \textbf{Energy ($E$)}: represents the overall system’s activity level. A sudden increase or decrease might indicate excessive vibrational wear or mechanical failure, which is crucial for RUL estimation.
    \begin{equation}
       E = \sum_{m=1}^{M} \left| \Gamma_{iw}(a,b) \right|^2
       \label{eq:energy}
    \end{equation}

    \item \textbf{Dominant frequency (\(f_d \))} represents the frequency at which \(E \) is the highest, revealing developing flaws, including varying defects, misalignment, or imbalance in rotating equipment.
    \begin{equation}
        f_{d} = a_{\text{scale}} \left[ \arg\max(E) \right]
        \label{eq:fd}
    \end{equation}

    \item \textbf{Entropy $(h)$}: assesses the system's randomness and unpredictability. High entropy frequently indicates complex or chaotic behavior, which might be linked to non-stationarynonstationary defects or irregular operational circumstances.
    \begin{equation}
       h = -\sum_{i=m}^{M} P(i_w(t)) \log P(i_w(t))
       \label{eq:entropy}
    \end{equation}

    \item \textbf{Kurtosis $(K)$}: measures the "tailedness" of the signal distribution, which helps in detecting extreme variations.
    \begin{equation}
        K = \frac{\mathbb{E}[(i_w(t) - \mu)^4]}{\sigma^4}
        \label{eq:kurtosis}
    \end{equation}

    \item \textbf{Skewness $(Sk)$} represents the asymmetry of the signal distribution. Significant positive or negative skewness can indicate uneven operational stress or asymmetric degradation in components.
    \begin{equation}
        s_k = \frac{\mathbb{E}[(i_w(t) - \mu)^3]}{ \sigma^3}
        \label{eq:skewness}
    \end{equation}

    \item \textbf{Mean}: The mean value provides the baseline operating condition of the system. Deviation from the expected mean may indicate shifts in operational parameters or degradation.
    \begin{equation}
        \text{avg} = \frac{1}{M} \sum_{m=1}^{M} i_w(m)
        \label{eq:mean}
    \end{equation}

    \item \textbf{Standard deviation (\(\sigma_{TF}\))}: Indicates high variability in the signal, often linked to unstable operation or the presence of faults.
    \begin{equation}
     \sigma_{TF} = \sqrt{\frac{1}{M} \sum_{i=1}^{M} \left(i_w(m) - \mu \right)^2}
     \label{eq:std}
    \end{equation}
\end{itemize}

\begin{algorithm}[htbp]
\caption{Time‒frequency algorithm}\label{algo:tf}
\footnotesize
\begin{algorithmic}[1]
\Require vertical-channel windowed segments ($I_{wv}$), critical frequency ($f_c$), operating frequency ($f_o$), and sampling frequency ($f_{sampling}$)
\State $a_{min} = \frac{f_c}{f_{max}\cdot f_{sampling}}, \quad a_{{max}} = \frac{f_c}{f_{\text{min}}\cdot f_{sampling}}$
\State $a_{scale} \in [a_{min}, a_{max}]$
\State $I_{{TF}_{v}} \gets \{\}$

\For{$i_w$ in $I_{wv}$:}
        \State Compute $\Gamma_{i_w}(a,b)$ as in Equation \ref{eq:wavelet}.
        \State Compute $E_v$ as in Equation \ref{eq:energy}.
        \State Compute $f_{d_{h}}$ as in Equation \ref{eq:fd}.
        \State Compute $h_v$ as in Equation \ref{eq:entropy}.
        \State Compute $K_v $ as in Equation \ref{eq:kurtosis}.
        \State Compute $sk_v$ as in Equation \ref{eq:skewness}.
        \State Compute $\text{avg}_v$ as in Equation \ref{eq:mean}.
        \State Compute $\sigma_{TF_v} $ as in Equation \ref{eq:std}.
        \State $I_{{TF}_{v}} \gets [\log(E_v), f_{d_{v}}, h_v, K_v, sk_v, \text{avg}_v, \sigma_{TF_v}] $
\EndFor
\State \textbf{return} $I_{{TF}_{v}}$
\end{algorithmic}
\end{algorithm}

\subsubsection{Multimodal Sequential Data Generator}
\begin{figure}[t]
\centering
\includegraphics[width=1.0\textwidth]{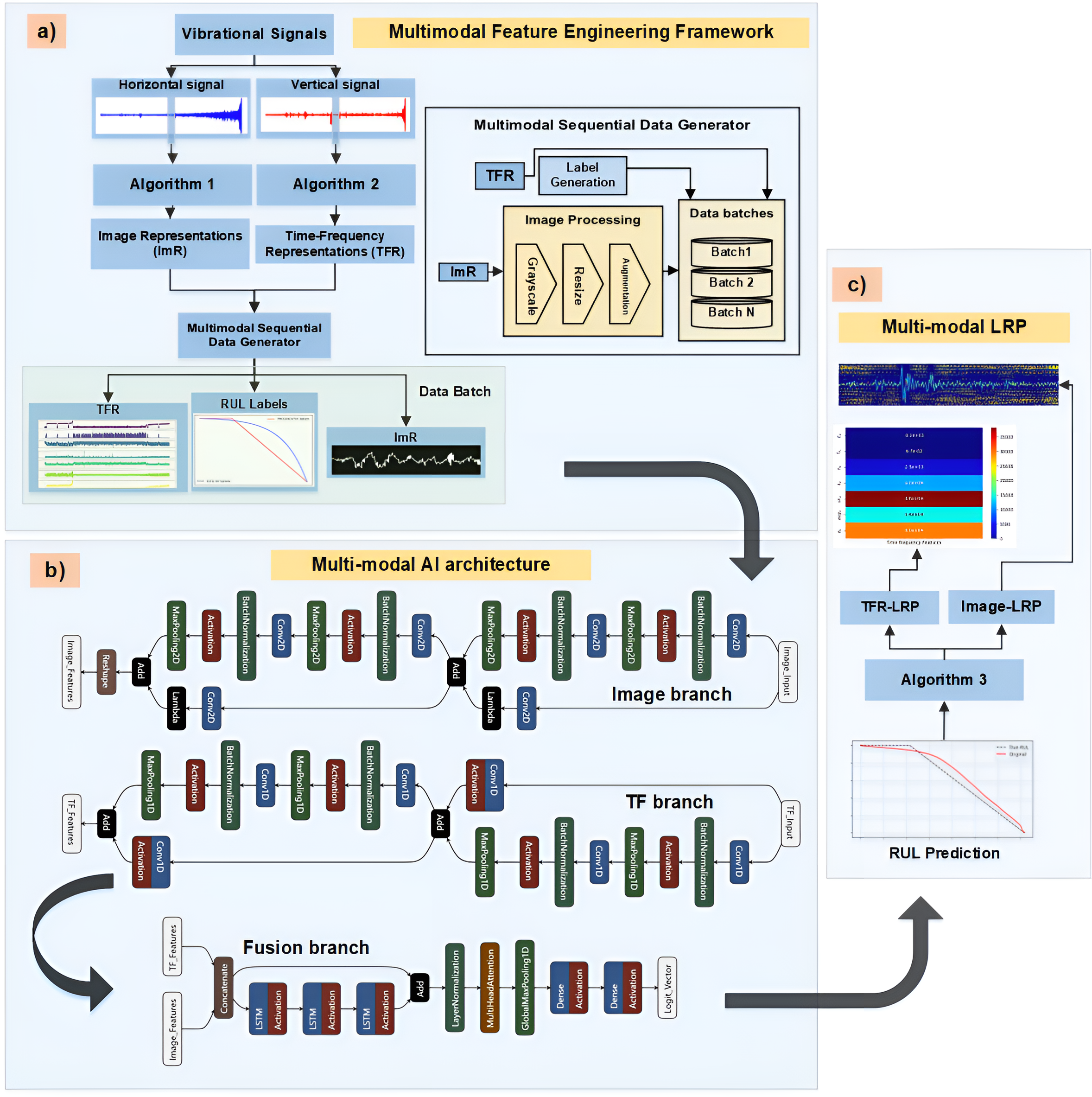}
\caption{Visual representation of the proposed multimodal RUL framework. \textbf{(a)} Horizontal and vertical vibrational signals are rectangularized and processed using Algorithm~\ref{algo:rast} and Algorithm~\ref{algo:tf} to generate the ImR and TFR, respectively. These are then fed into the Multimodal Sequential Data Generator to apply image processing techniques and generate RUL labels. \textbf{(b)} A multimodal AI model with parallel Image and TF branches processes the data, and their outputs are fused in fusion branch to estimate RUL. \textbf{(b)} The estimated RUL is propagated backward using Algorithm~\ref{algo:lrp} for Multimodal-LRP explanations.}
\label{fig:model}
\end{figure}

We developed a multimodal sequential data generator to streamline the training process and manage computational complexity. The generator takes both ImR and TFR data. For the image, we implemented some image processing techniques to ease the computational requirements. The images are first converted into grayscale and then resized to (\(64\times500 \)), and then random augmentations (e.g., flipping, noise, etc.) are applied to increase the diversity in the data to increase the model’s ability to handle dynamic situations. The TF data are utilized to generate RUL labels from training, and then the total data, including labels, are divided sequentially into smaller batches for efficient training. The graphical illustration is shown in Figure \ref{fig:model}(a).

\subsection{Multimodal-AI Network}
We propose a novel multimodal-AI network to estimate the RUL of rolling-element bearings by leveraging ImR and TFR. Figure \ref{fig:model}(b) shows an illustration of the network. The framework comprises three distinct branches described as:

\subsubsection{Image Branch}
The image branch processes ImR and extracts spatial degradation patterns via a series of convolutional blocks. In our implementation, the main path comprises two convolutional blocks, each consisting of multiple dilated convolutional layers. The first block employs smaller filter sizes with a high dilation rate, enabling the model to capture intricate local features while preserving a large receptive field, improving the detection of subtle degradation cues that may otherwise overlooked. In contrast, the second block consists of larger filter sizes with lower dilation rates, focusing on capturing large-scale degradation patterns and extracting global spatial correlations, crucial for accurate RUL estimation. To improve training stability and computational efficiency, residual connections are incorporated at the end of each block essentially forming a residual path that preserves identity mapping and enhances gradient flow during backpropagation. The hyperparameters of the image branch are provided in Table \ref{tab:hp_implement}.

\begin{table}[htbp]
    \caption{Hyperparameters of our implementation}
    \label{tab:hp_implement}
    \centering
    \footnotesize
    \begin{tabular}{@{}lcc@{}}
        \toprule
        \textbf{Branch} & \textbf{Main Path} & \textbf{Residual Path} \\
        \midrule

        \multirow{2}{*}{\textbf{Image}} 
        & \makecell[l]{
            Conv2D Layers: 4 \\
            Filters: [32, 32, 64, 64] \\
            Kernel Sizes: [(5,5), (3,3), (3,3), (2,2)] \\
            Dilation: [(4,4), (3,3), (2,2), (1,1)] \\
            Padding: Same \\
            Regularization Strength: 0.01 \\
            Activation: ReLU \\
            MaxPooling2D: [(3,3), (3,3), (3,3), (2,2)] \\
            Reshape: Match shapes
        } 
        & \makecell[l]{
            Conv2D Layers: 2 \\
            Filters: [32, 64] \\
            Kernel Sizes: [(1,1), (1,1)] \\
            Regularization Strength: 0.01 \\
            Lambda Layer: Resizing \\
        } \\
        \midrule

        \multirow{2}{*}{\textbf{TF}} 
        & \makecell[l]{
            Conv1D Layers: 4 \\
            Filters: [32, 32, 64, 64] \\
            Kernel Sizes: [2, 2, 2, 2] \\
            Dilation: [2, 2, 1, 1] \\
            Padding: Same \\
            Regularization Strength: 0.01 \\
            Activation: ReLU \\
            MaxPooling1D: [1, 1, 1, 1]
        } 
        & \makecell[l]{
            Conv1D Layers: 2 \\
            Filters: [32, 64] \\
            Kernel Sizes: [1, 1] \\
            Regularization Strength: 0.01 \\
        } \\
        \midrule

        \textbf{Fusion} 
        & \multicolumn{2}{l}{\makecell[l]{
            Concatenate Axis: 0 \\
            LSTM Layers: 3, Units: [100, 64, 64], Activation: tanh \\
            Residual Add: After LSTM block \\
            Regularization Strength: 0.01, LayerNorm: Yes \\
            Multi-Head Attention: 8 Heads (Key, Dim=64) \\
            Linear Layers: 3, Units: [64, 32, 1]
        }} \\
        \bottomrule
    \end{tabular}
\end{table}

\subsubsection{TF Branch}
The TF branch processes the extracted TFR by passing it through a structure similar to the image branch. By leveraging dilation in convolutional layers, the module expands its receptive fields, allowing it to detect subtle degradation patterns while preserving the ability to focus on localized signal variations. This capability ensures that the model can accurately estimate the RUL by effectively learning both short-term fluctuations and large-scale degradation patterns in the TFR. The hyperparameters of the TF branch are provided in Table \ref{tab:hp_implement}.

\subsubsection{Fusion Branch}
The fusion branch receives the output from both the image and TF branches and combines them through concatenation, which is then processed by the LSTM block to capture combined long-term time dependencies with added stability and facilitate identity mapping through residual connections. The output of the LSTM block is passed through the MHA mechanism to filter out relevant, useful information while suppressing redundancy crucial. The filtered features are fed into a series of linear layers, where patterns are refined and the final RUL is predicted. The hyperparameters of the fusion branch are provided in Table \ref{tab:hp_implement}.

\subsection{Multimodal-LRP}
We propose a novel Multimodal-LRP algorithm to enhance interpretability within our multimodal RUL framework by extending LRP to neural networks that process multiple input modalities. Traditional LRP approaches \cite{montavon2019layer, bach2015pixel, samek2016interpreting} typically assume linear or sequential architectures, limiting their applicability in multimodal settings. Our method, detailed in Algorithm~\ref{algo:lrp}, explicitly addresses this challenge by introducing structured tracking mechanisms during both the forward and backward passes. During the \textit{forward pass}, activations are computed sequentially for each layer in the network, with special handling of \texttt{Add} (residual) operations and \texttt{Concatenate} layers. When an \texttt{Add} operation is encountered, the algorithm identifies and stores the residual input activations using the \texttt{FindResidualInput} procedure. Similarly, when a \texttt{Concatenate} layer is encountered, the combined activations are carefully recorded. This structured tracking ensures that all multimodal interactions and shortcut pathways are accurately preserved for interpretability. In the \textit{backward pass}, these recorded operations play a crucial role in \textit{relevance propagation}. For residual (\texttt{Add}) layers, the algorithm splits the relevance score into two distinct contributions: one for the main path and one for the residual path. The stored residual activations are used to correctly attribute the relevance flowing through the skip connection, avoiding dilution or over-attribution. In the case of \texttt{Concatenate} operations, the backward pass relies on the ordering and structure of the forward-tracked activations to appropriately disentangle the relevance contributions to each original modality (ImR vs. TFR). This ensures that the resulting relevance maps are modality-specific, accurately reflecting how each input influenced the model’s output. To further clarify the relevance score redistribution, we also present a detailed mathematical explanation for each layer to highlights the contribution of each layer to the overall prediction process.

\begin{algorithm}
\caption{Multimodal-LRP algorithm}\label{algo:lrp}
\begin{algorithmic}[1]

\Require Input $x$, Neural Network $f(. )$

\Procedure{Forward}{$x, f$}
    \State Initialize activations $A \gets [x]$, residual outputs $R_{res} \gets \emptyset$

    \For{layer in $f.layers$}
        \If{layer is Add}
            \State $residual \gets R_{res}.get(\text{FindResidualInput}(layer), A[-1])$
            \State $A_{new} \gets layer.forward([A[-1], residual])$
            \If{residual exists} $R_{res}[layer] \gets A_{new}$ \EndIf
        \ElsIf{layer is Concatenate}
            \State $A_{new} \gets layer.forward(A)$
        \Else
            \State $A_{new} \gets layer.forward(A[-1])$
        \EndIf
        \State Append $A_{new}$ to $A$
    \EndFor

    \State \textbf{Return:} $A_{out}$
\EndProcedure

\Procedure{Backward}{$R^{(logit)}, f$}
    \State Initialize relevance $R^{(l)} \gets R^{(logit)}$

    \For{layer in reversed($f.layers$)}
        \If{layer is Add}
            \State $(R_{main}, R_{residual}) \gets layer.backward(R^{(l)})$
            \State $R^{(l)} \gets R_{main} + R_{residual} + R_{res}.get(layer, 0)$
        \Else
            \State $R^{(l)} \gets layer.backward(R^{(l)})$
        \EndIf
    \EndFor

    \State \textbf{Return:} Relevance maps $R_{img}, R_{tf}$
\EndProcedure

\Procedure{FindResidualInput}{layer}
    \State \textbf{Return:} residual connections mapped inputs
\EndProcedure

\end{algorithmic}
\end{algorithm}

\subsubsection{Convolutional Layer}

The convolutional layer in both the image and TF branches computes the distribution via LRP-$\gamma$, with more weights given to positive weights, as follows:

\begin{equation*}
    w_{\text{conv},\mathbf{m}, c}^{-} = \min(w_{\text{conv},\mathbf{m}, c}, 0) \quad \text{and} \quad w_{\text{conv},\mathbf{m}, c}^{+} = \max(w_{\text{conv},\mathbf{m}, c}, 0)
\end{equation*}

The contributions are then calculated as follows:

\begin{equation*}
    z_{\text{conv},\mathbf{i}}^{+} = \sum_{\mathbf{m}, c} w_{\text{conv},\mathbf{m}, c}^{+} \cdot x_{\text{conv},\mathbf{i} + \mathbf{m}, c} \quad \text{and} \quad z_{\text{conv},\mathbf{i}}^{-} = \sum_{\mathbf{m}, c} w_{\text{conv},\mathbf{m}, c}^{-} \cdot x_{\text{conv}}
\end{equation*}

The sensitivity parameter is computed as:

\begin{equation*}
    S_{\text{conv},\mathbf{i}} = \frac{R_{\mathbf{i}}^{(l+1)}}{(1 + \gamma) z_{\text{conv},\mathbf{i}}^{+} - z_{\text{conv},\mathbf{i}}^{-}}
\end{equation*}

The final relevance distribution is computed as:

\begin{equation}
     S_{\text{conv},,\mathbf{i}} \cdot w_{\text{adjusted},\text{conv},\mathbf{m}, c}
    \label{eq:conv}
\end{equation}

where \(w_{\text{adjusted}, \text{conv}, \mathbf{m}, c} \) is the adjusted weight computed as:

\begin{equation}
    w_{\text{adjusted},\text{conv}, \mathbf{m}, c} = (1 + \gamma) w_{\text{conv},\mathbf{m}, c}^{+} + w_{\text{conv},\mathbf{m}, c}^{-}
\end{equation}

\subsubsection{Linear Layers}
The relevance of neuron \(i \) in linear layer \(l \) is computed on the basis of its contribution \(C_{\text{dense}, i}^{(l)} \) through LRP-$\epsilon$ as:

\begin{equation}
    R_{\text{dense}, i}^{(l)} = C_{\text{dense}, i}^{(l)} \cdot x_{\text{dense}, i}^{(l)}
    \label{eq:linear}
\end{equation}

where the contribution is given by:

\begin{equation}
    C_{\text{dense}, i}^{(l)} = \sum_j w_{\text{dense}, ij}^{(l)} \cdot S_{\text{dense}, j}^{(l)}
\end{equation}

and the sensitivity parameter is defined as:

\begin{equation}
    S_{\text{dense}, j}^{(l)} = \frac{R_{\text{dense}, j}^{(l+1)}}{z_{\text{dense}, j}^{(l)} }
\end{equation}

where preactivation value \(z_{\text{dense}, j}^{(l)} \) for each neuron \(j \) in layer \(l \) is defined as:

\begin{equation}
    z_{\text{dense}, j}^{(l)} = \sum_i x_{\text{dense}, i}^{(l)} \cdot w_{\text{dense}, ij}^{(l)} + b_{\text{dense}, j}^{(l)}
\end{equation}

\subsubsection{MaxPooling}
The relevance distribution through MaxPooling layers is computed via LRP-0 as:

\begin{equation}
     R_{Pool,i} =
\begin{cases}
{R_{j}^{(l)}}/{N_{i}} & \text{if } x_{i, t} = \max_{t'} x_{i, t'} \\
0 & \text{otherwise}
\end{cases}
\label{eq:pool}
\end{equation}

where \(R_{i}^{(l)} \) is the relevance of neuron \(i \) and \(N_i \) is the number of time steps where multiple time steps have the same maximum values, ensuring that the relevance is conserved across layers.

\subsection{Multi-head Attention Network}
The multi-head attention network is a complex architecture that requires detailed backward derivation of relevance propagation, which is beyond the scope of this research. We computed a distributed relevance score based on LRP-0 via the normalized attention weights \(A_{i,h,t,t'} \) as follows:
\begin{equation}
    R_{\text{mha}, i} = \sum_{h,t'} \frac{A_{i, h, t, t'}}{\sum_{t''} A_{i, h, t', t''}} R_{j}^{(l+1)}
   \label{eq:mha}
\end{equation}

\subsubsection{Normalization}
We utilized LRP-$\epsilon$ to distribute the relevance in each normalization layer as:
\begin{equation}
    R_{norm,i}^{(l)} = \frac{\gamma_f \cdot R_{j}^{(l+1)}}{\sqrt{\sigma_{i}^2 + \epsilon}},
    \label{eq:ln}
\end{equation}
where $\gamma_f$ is the scaling factor in the forward pass and
\begin{equation}
    \sigma_{i} = \frac{1}{F} \sum_{f=1}^{F} \left(x_{i,f} - \mu_i \right)^2,
    \label{eq:variance}
\end{equation}
and
\begin{equation}
    \mu_i = \frac{1}{F} \sum_{f=1}^{F} x_{i,f}.
\end{equation}

\subsubsection{Add layer}
The residual connection (Add) can be redistributed on the basis of inputs via the LRP-0 rule as follows:
\begin{equation}
    [R_{\text{add}, i}^{(input1)}, R_{\text{add}, i}^{(input2)}] = \frac{R_{j}^{(l+1)}}{N}
    \label{eq:add}
\end{equation}
where \(N \) is the number of inputs, usually two in the residual connection.

\subsection{LSTM}
Similar to Multi-head attention, the LSTM architecture is complex and requires detailed backward propagation, which exceeds the scope of this research. The relevance is computed using LRP-0 as follows:
\begin{equation}
    R_{\text{lstm}, i}^{(l)} = \sum_{h} \frac{|w^{(l)}_{f,h}|}{\sum_{f} |w^{(l)}_{f,h}|} \cdot R_{j}^{(l+1)}
    \label{eq:lstm}
\end{equation}
where \(w_{f,h}^{(l)} \) are the weights connecting the \(f \)-th input feature to the \(h \)-th hidden unit, and \(R_{\text{lstm}, j}^{(l+1)} \) is the relevance of the preceding layer \(l+1 \).

\subsubsection{Concat}
The concat node receives two relevances, i.e., \(R_{\text{lstm}, i}^{(1)} \) and \(R_{\text{add}, i}^{(\text{concat})} \). First, the relevance is averaged as follows:
\begin{equation}
     R_{\text{concat}, i} = \frac{R_{\text{lstm}, i}^{(1)} + R_{\text{add}, i}^{(concat)}}{2}
    \label{eq:concat}
\end{equation}
then the relevances are distributed using the \texttt{split(.)} function based on the corresponding input tensor image or TF using LRP-0 as follows:
\begin{equation}
    R_{\text{img}, i}, R_{\text{TF}, i} = \texttt{split}(R_{\text{concat}, i},N)
\end{equation}

\subsubsection{Reshape Layer}
A Reshape layer is introduced at the output of the image branch to make the shapes compatible with the TF branch for the concatenation process. \(R_{\text{img}, i} \) is propagated backward to match the shape of the corresponding block using \texttt{Reshape}(.) and computed as LRP-0 as follows:
\begin{equation}
R_{\text{reshape}, i} = \texttt{Reshape}(R_{\text{img}, j})
\label{eq:reshape}
\end{equation}

\subsubsection{Lambda Layer}
The lambda layer in the image branch ensures that the shape is compatible with residual addition computed via LRP-0 as:
\begin{equation}
   R_{\lambda, i}^{(l)} = \texttt{resize}(x_{i}^{(l)}) \cdot R_{j}^{(l+1)}
   \label{eq:lambda}
\end{equation}

\subsubsection{Activation}
The relevance distribution of the activation layer is computed using basic LRP-0 as follows:
\begin{equation}
   R_{\text{actv}, i}^{(l)} = R_{j}^{(l+1)} \cdot \max(0, x_i^{(l)})
   \label{eq:actv}
\end{equation}

\section{Data Analysis, Evaluation Metrics, and Training Optimizations} \label{sec:data_exp}
This section overviews the data utilized for training, evaluation metrics, and training and testing optimization used to mitigate computational overhead.

\subsection{Dataset Explanation}
\begin{figure}[h!]
\centering
\includegraphics[width=0.9\textwidth]{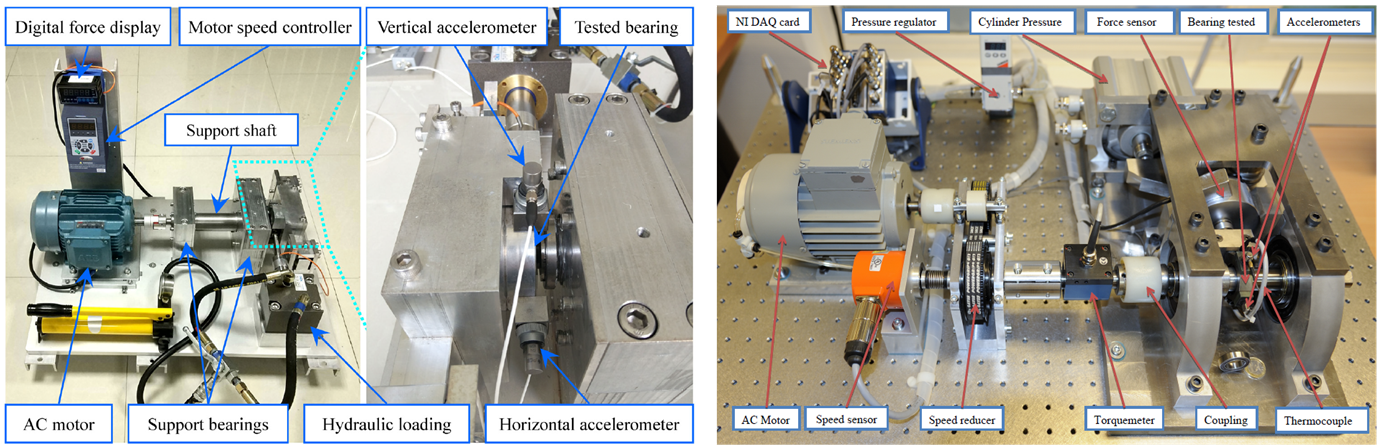}
\caption{\textbf{a)} XJTU-SY testbed; \textbf{b)} PRONOSTIA testbed for recording vibrational data}
\label{fig:testbed}
\end{figure}

\textbf{XJTU-SY Dataset:} is a benchmark dataset for RUL estimation of rolling-element bearings developed by Xian Jiaotong in collaboration with Chanxing Sumyoung Technology \cite{wang2018hybrid}. The dataset contains complete run-to-failure recordings for 15 rolling-element bearings, collected during accelerated degradation experiments conducted on the testbed shown in Figure \ref{fig:testbed}(a). Data were acquired under three distinct operating conditions: 1200 rpm (35 Hz) with 12 kN radial load; 2250 rpm (37.5 Hz) with 11 kN radial load; 2400 rpm (40 Hz) with 10 kN radial load. Vibrational signals were recorded using accelerometers mounted on horizontal and vertical axes, sampled and 25.6 kHz. Data were logged at one-minute intervals, with each sample capturing 1.28 seconds of signal data. For training, we selected three bearing datasets (Bearing 3$\sim$5) corresponding to the 35 Hz condition, which accounts for approximately 72$\%$ of the data under this condition. Horizontal axis signals were converted into ImR using the Algorithm \ref{algo:rast}, while vertical axis signals were processed to extract TFR using the Algorithm \ref{algo:tf}. The data from the 37.5 Hz operational condition was used to evaluate the generalization capability of the trained AI model. The detailed training and testing splitting of data is provided in the Table \ref{tab:data_exp}. The 40 Hz condition was excluded from testing due to computational constraints associated with our system setup.

\textbf{PRONOSTIA Dataset:} is a widely recognized benchmark used in research focused on condition monitoring and RUL prediction of rolling-element bearings. It was developed as part of the PRONOSTIA experimental platform \cite{nectoux2012pronostia}. The dataset comprises 17 complete run-to-failure vibrational signals, collected under accelerated degradation conditions across three operational conditions: 1800 rpm (100 Hz) with 4 kN radial load; 1600 rpm (100 Hz) with 4.2 kN load; 1500 rpm (100 Hz) with 5 kN load. Vibration signals were recorded using accelerometers mounted along the horizontal and vertical axes (see Figure \ref{fig:testbed}(b)), with a sampling rate of 25.6 kHz, and temperature data was recorded at 10 Hz. For training, we selected data from three bearings (Bearing 4$\sim$7) operating under the 4 kN load condition, representing approximately 52$\%$ of the total data available for that specific operational setting. As with XJTU-SY, horizontal vibration signals were converted into ImR using the Algorithm \ref{algo:rast}, while vertical axis signals were processed to extract TFR using the Algorithm \ref{algo:tf}. The data from other operating conditions was used to test the generalization of the trained AI. The detailed training and testing splitting of data is provided in the Table \ref{tab:data_exp}. The temperature data was ignored during training and testing.

\begin{table}[htbp]
    \caption{Detailed working conditions in datasets XJTU-SY and PRONOSTIA}
    \vspace{3mm}
    \label{tab:data_exp}
    \centering
    \footnotesize
    \begin{tabular}{@{}lccccccc@{}}
        \toprule
        \textbf{Dataset} & \textbf{Condition} & \textbf{Frequency} & \textbf{Loading Force} & \textbf{Rotating Speed} & \textbf{Total Bearings} & \textbf{Train Data} & \textbf{Test Data} \\
        \midrule
        \multirow{3}{*}{XJTU}
        & Condition1 & 35 Hz & 12 kN & 2100 rpm & 5 & 3$\sim$5 & 1$\sim$2\\
        & Condition2 & 37.5 Hz & 11 kN & 2250 rpm & 5 &\text{-} &All\\
        & Condition3 & 40 Hz & 10 kN & 2400 rpm & 5 &\text{-}&\text{-}\\
        \midrule
        \multirow{3}{*}{PRONOSTIA}
        & Condition1 & 100 Hz & 4 kN & 1800 rpm & 7 & 4$\sim$7 &1$\sim$3\\
        & Condition2 & 100 Hz & 4.2 kN & 1650 rpm & 7 &\text{-}& All \\
        & Condition3 & 100 Hz & 5 kN & 1500 rpm & 3 &\text{-} &All \\
        \bottomrule
    \end{tabular}
\end{table}

\subsection{RUL Labels}
\begin{figure}[ht!]
\centering
\includegraphics[width=0.5\textwidth]{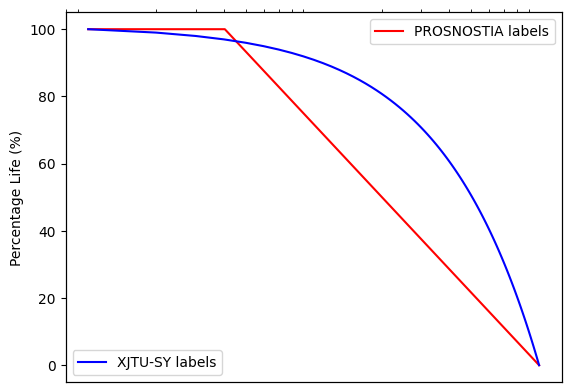}
\caption{RUL labels for both datasets.}
\label{fig:labels}
\end{figure}

Accurate RUL label distribution is a critical step in developing effective RUL estimation models. While some studies assume a constant-rate degradation, real-world scenarios often exhibit nonlinear and irregular degradation patterns. To explore both perspectives, we generated RUL labels for the XJTU-SY dataset using a linear degradation model and visualized on a logarithmic scale in Figure \ref{fig:labels} for better visualization. Since the long-term monitoring data naturally form a time series relation, the initial phase of bearing operating typically reflects a stable condition with minimal observable degradation. To better capture these characteristics, we utilized a nonlinear piecewise degradation distribution for the PRONOSTIA dataset as illustrated In Figure \ref{fig:labels}. The approach represents the gradual decline in bearing performance over time more realistically.

\subsection{Evaluation Metrics}

To assess the training and testing performance of our architecture, we utilized two commonly used evaluation metrics for regression tasks:

\textbf{Mean Absolute Error (MAE)}: a widely used metric in RUL analysis to quantify the accuracy of predictions. It measures the average magnitude of the absolute error between the predicted RUL ($\hat{y}_i$) and the ground truth RUL ($y_i$) irrespective of the direction.
\begin{equation}
    \text{MAE} = \frac{1}{n} \sum_{i=1}^{n} \left| y_i - \hat{y}_i \right|
\end{equation}
where \(n\) represents the total number of predictions. The MAE is particularly suitable for RUL estimation because it ensures an unbiased evaluation of prediction accuracy by penalizing both overestimations and underestimations equally.

\textbf{Mean Square Error (MSE)}: calculates the square root of the average squared difference between the predicted $\hat{y}_i$ and the ground truth $y_i$:
\begin{equation}
    \text{MSE} = \sqrt{\frac{1}{n} \sum_{i=1}^{n} \left(y_i - \hat{y}_i \right)^2}
\end{equation}
The MSE penalizes larger errors more heavily due to the squaring of differences, making it more sensitive to significant deviations in prediction error. It emphasizes the model’s ability to minimize large prediction errors, making it useful for scenarios where extreme misestimations must be avoided.

\subsection{Training Optimizations}\label{sec:train_opt}

To optimize the training process and address computational bottlenecks, we implemented several regularization techniques to achieve the best training results described below:

\subsubsection{Training Callbacks}

To ensure that the model converges to the best solution, we incorporate several callbacks: \texttt{ResetStateCallback} to reset model states between epochs, \texttt{EarlyStopping} to stop training if validation loss stagnates after a set number of epochs, \texttt{ReduceLROnPlateau} to adjust the learning rate dynamically based on MSE loss, and \texttt{ModelCheckpoint} to save the best training weights, collectively improving both training efficiency and model performance.

\subsubsection{K-fold Cross Validation and Training}

We utilized k-fold cross-validation with \(K\) folds to enhance the model’s robustness. Each fold is partitioned into training and validation sets using indices generated by k-fold splits. The use of k-folds reduces overfitting, improves generalization, and offers a reliable framework for hyperparameter tuning and unbiased validation, ensuring that the model performs well across a range of operating conditions.

\subsection{Overall work flow}
The overall workflow for the multimodal RUL framework can be summarized in these steps:
\begin{itemize}
    \item Capture the vibrational signals from a multichannel accelerometer and segment the signal using Equation \ref{eq:window}.
    \item Normalize the horizontal signal using Equation \ref{eq:minmax}, and convert it into an ImR using Algorithm \ref{algo:rast}.
      \item Split the data into training and testing fold as described in Table~\ref{tab:data_exp}.
    \item Train the multimodal-AI network using the training optimizations outlined in Section~\ref{sec:train_opt}.
    \item To predict the RUL, implement the \textit{forward pass} procedure described in the Algorithm \ref{algo:lrp} during inference to store the activations from \texttt{Add} and \texttt{Concatenation} operations.
    \item To compute multimodal-LRP, utilize the stored activations from the \texttt{Add} and \texttt{Concatenation} operations during the \textit{backward pass} to accurately redistribute the relevance score as outlined in Algorithm \ref{algo:lrp}.
\end{itemize}

\begin{figure}[htbp]
\centering
\includegraphics[width=0.8\textwidth]{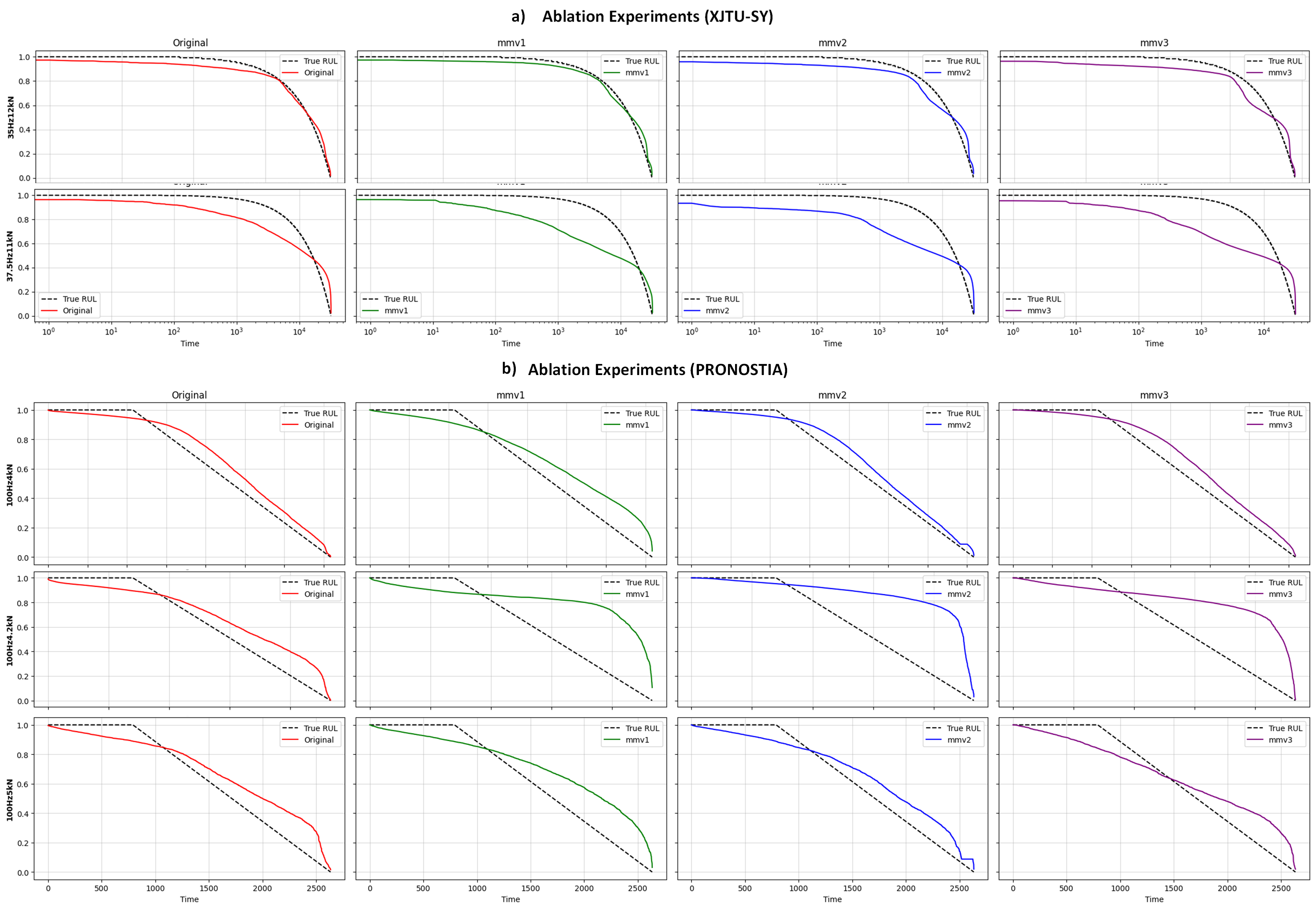}
\caption{Ablation experiment results for \textbf{a)} XJTU-SY; \textbf{b)} PRONOSTIA.}
\label{fig:ablation}
\end{figure}

\section{Experiments and Validations}\label{sec:results}
In this section, we analyze the generalizability, robustness and performance comparison with baseline models of our proposed framework by subjecting it to different experiments.

\subsection{Ablation Experiments}
To assess the effectiveness of each component in our proposed framework, we conducted ablation experiments. The original framework was modified to create three variants: MMv1 (without dilation in the convolutional network), MMv2 (without residual connections), and MMv3 (without the attention mechanism). Based on the MSE and MAE metrics presented in Table~\ref{tab:ablation_xjtu_sy} for the XJTU-SY dataset and Table~\ref{tab:ablation_pronostia} for the PRONOSTIA dataset, the original model consistently outperforms all ablated versions, achieving the lowest error values and demonstrating strong generalization to unseen conditions. Among the ablated models, MMv2 ranks second with slightly higher error values; in some cases, it even surpasses the original model. MMv1 performs relatively poorly in generalized scenarios underlines the importance of dilation. MMv3 ranks fourth overall, which underlines the importance of the attention mechanism.

To further interpret the results and explain generalization, we selected Bearing 1 from each dataset and condition for detailed analysis. For the XJTU-SY dataset, results for each condition are visualized in Figure~\ref{fig:ablation}(a). In condition 1, the original model achieves the best performance with an MSE of 0.00231 and an MAE of 0.04086. MMv1 follows with an MSE of 0.00281 and an MAE of 0.04653. MMv3 comes next with an MSE of 0.00293 and an MAE of 0.04799, while MMv2 ranks fourth with an MSE of 0.00296 and an MAE of 0.04855. A similar trend is observed in condition 2, where the original model again outperforms all variants with an MSE of 0.02006 and an MAE of 0.12796. MMv1 secures the second position with an MSE of 0.03098 and an MAE of 0.14791, followed by MMv2 with an MSE of 0.03474 and an MAE of 0.16431. MMv3 performs the worst in this condition, with an MSE of 0.03804 and an MAE of 0.17113.

For the PRONOSTIA dataset, the results across all conditions are presented in Figure~\ref{fig:ablation}. In condition 1, MMv2 slightly outperforms the original model, achieving an MSE of 0.00104 and an MAE of 0.02308, while the original model follows closely with an MSE of 0.00148 and an MAE of 0.03280. MMv3 records an MSE of 0.00151 and an MAE of 0.03366, and MMv1 performs the worst, with an MSE of 0.00931 and an MAE of 0.08179. In condition 2, the original model returns to the top position, achieving an MSE of 0.01003 and an MAE of 0.08686. MMv3 comes second with an MSE of 0.06056 and an MAE of 0.18506, followed by MMv1 with an MSE of 0.06668 and an MAE of 0.20196. MMv2 ranks fourth in this condition, with an MSE of 0.07983 and an MAE of 0.19818. In condition 3, a similar pattern to condition 1 is observed, where MMv2 performs best with an MSE of 0.00725 and an MAE of 0.07415. The original model ranks second with an MSE of 0.01021 and an MAE of 0.08618. MMv1 and MMv3 follow with MSE values of 0.01654 and 0.01673, and MAE values of 0.11047 and 0.11159, respectively.

\begin{table}[htbp]
\centering
\caption{Ablation experiment (XJTU-SY)}
\vspace{2mm}
\label{tab:ablation_xjtu_sy}
\footnotesize
\setlength{\tabcolsep}{5pt}
\renewcommand{\arraystretch}{0.6}
\begin{tabular}{@{}llSSSS@{}}
\toprule
\rowcolor{gray!20}
\multirow{2}{*}{\textbf{Bearing}} & \multirow{2}{*}{\textbf{Model}} &
\multicolumn{2}{c}{\textbf{35Hz12kN}} &
\multicolumn{2}{c}{\textbf{37.5Hz11kN}} \\
\cmidrule(lr){3-4} \cmidrule(lr){5-6}
 & & \textbf{MSE} & \textbf{MAE} & \textbf{MSE} & \textbf{MAE} \\
\midrule

\multirow{5}{*}{\textbf{Bearing 1}}
 & Original & \textbf{0.00231} & \textbf{0.04086} & \textbf{0.02006} & \textbf{0.12796} \\
 & MMv1 & 0.00281 & 0.04653 & 0.03098 & 0.14791 \\
 & MMv2 & 0.00296 & 0.04855 & 0.03474 & 0.16431 \\
 & MMv3 & 0.00293 & 0.04799 & 0.03804 & 0.17113 \\
 
\midrule

\multirow{5}{*}{\textbf{Bearing 2}}
 & Original & 0.00357 & 0.05182 & 0.03374 & 0.16286 \\
 & MMv1 & 0.00205 & 0.03502 & \textbf{0.01641} & \textbf{0.10939} \\
 & MMv2 & \textbf{0.00175} & \textbf{0.03193} & 0.02188 & 0.12963 \\
 & MMv3 & 0.00222 & 0.03683 & 0.02128 & 0.12776 \\
\midrule

\multirow{5}{*}{\textbf{Bearing 3}}
 & Original & 0.00559 & 0.06174 & \textbf{0.00510} & \textbf{0.06651} \\
 & MMv1 & \textbf{0.00055} & \textbf{0.02032} & 0.01265 & 0.10388 \\
 & MMv2 & 0.00059 & 0.02088 & 0.00943 & 0.08681 \\
 & MMv3 & 0.00077 & 0.02430 & 0.01612 & 0.11497 \\
\midrule

\multirow{5}{*}{\textbf{Bearing 4}}
 & Original & \textbf{0.02337} & \textbf{0.13552} & \textbf{0.01518} & \textbf{0.11001} \\
 & MMv1 & 0.05842 & 0.20256 & 0.02448 & 0.13440 \\
 & MMv2 & 0.10482 & 0.27930 & 0.02180 & 0.13014 \\
 & MMv3 & 0.04862 & 0.17882 & 0.01897 & 0.11314 \\
\midrule

\multirow{5}{*}{\textbf{Bearing 5}}
 & Original & \textbf{0.01039} & \textbf{0.08696} & 0.00915 & 0.07942 \\
 & MMv1 & 0.01204 & 0.09806 & 0.00719 & 0.07695 \\
 & MMv2 & 0.01584 & 0.09917 & \textbf{0.00597} & \textbf{0.06676} \\
 & MMv3 & 0.01842 & 0.11926 & 0.00891 & 0.07942 \\

\bottomrule
\end{tabular}
\begin{minipage}{\textwidth}
\footnotesize
\vspace{3mm}
\textbf{Note:} Bold values indicate the minimum MSE and MAE for each bearing-condition combination.
\end{minipage}
\end{table}

\begin{table}[htbp]
\centering
\caption{Ablation Experiment (PRONOSTIA)}
\vspace{2mm}
\label{tab:ablation_pronostia}
\footnotesize
\setlength{\tabcolsep}{5pt}
\renewcommand{\arraystretch}{0.6}
\begin{tabular}{@{}llSSSSSS@{}}
\toprule
\rowcolor{gray!20}
\multirow{2}{*}{\textbf{Bearing}} & \multirow{2}{*}{\textbf{Model}} &
\multicolumn{2}{c}{\textbf{100Hz4kN}} &
\multicolumn{2}{c}{\textbf{100Hz4.2kN}} &
\multicolumn{2}{c}{\textbf{100Hz5kN}} \\
\cmidrule(lr){3-4} \cmidrule(lr){5-6} \cmidrule(lr){7-8}
 & & \textbf{MSE} & \textbf{MAE} & \textbf{MSE} & \textbf{MAE} & \textbf{MSE} & \textbf{MAE} \\
\midrule

\multirow{5}{*}{\textbf{Bearing 1}}
 & Original & 0.00148 & 0.03280 & \textbf{0.01003} & \textbf{0.08686} & 0.01021 & 0.08618 \\
 & MMv1 & 0.00931 & 0.08179 & 0.06668 & 0.20196 & 0.01654 & 0.11047 \\
 & MMv2 & \textbf{0.00104} & \textbf{0.02308} & 0.07983 & 0.19818 & \textbf{0.00725} & \textbf{0.07415} \\
 & MMv3 & 0.00151 & 0.03366 & 0.06056 & 0.18506 & 0.01673 & 0.11159 \\
 
\midrule

\multirow{5}{*}{\textbf{Bearing 2}}
 & Original & \textbf{0.00755} & \textbf{0.07163} & \textbf{0.00859} & \textbf{0.07791} & \textbf{0.01184} & \textbf{0.09474} \\
 & MMv1 & 0.01357 & 0.10114 & 0.04735 & 0.16662 & 0.07071 & 0.20583 \\
 & MMv2 & 0.01186 & 0.09011 & 0.05047 & 0.15706 & 0.07646 & 0.19557 \\
 & MMv3 & 0.01126 & 0.08977 & 0.04554 & 0.15393 & 0.05526 & 0.17760 \\
\midrule

\multirow{5}{*}{\textbf{Bearing 3}}
 & Original & \textbf{0.00357} & \textbf{0.05222} & \textbf{0.01174} & \textbf{0.09351} & 0.03588 & 0.17025 \\
 & MMv1 & 0.00447 & 0.05687 & 0.08591 & 0.21200 & 0.05100 & 0.16999 \\
 & MMv2 & 0.00280 & 0.04534 & 0.06282 & 0.18750 & 0.06148 & 0.17758 \\
 & MMv3 & 0.00228 & 0.04015 & 0.05265 & 0.17061 & \textbf{0.04640} & \textbf{0.15699} \\
\midrule

\multirow{5}{*}{\textbf{Bearing 4}}
 & Original & \textbf{0.00259} & \textbf{0.03638} & \textbf{0.00929} & \textbf{0.08340} & \text{--} & \text{--} \\
 & MMv1 & 0.01373 & 0.10340 & 0.05910 & 0.18125 & \text{--} & \text{--} \\
 & MMv2 & 0.01276 & 0.09151 & 0.07538 & 0.19658 & \text{--} & \text{--} \\
 & MMv3 & 0.00780 & 0.06219 & 0.05898 & 0.17495 & \text{--} & \text{--} \\
\midrule

\multirow{5}{*}{\textbf{Bearing 5}}
 & Original & \textbf{0.04703} & \textbf{0.18604} & \textbf{0.06076} & \textbf{0.20063} & \text{--} & \text{--} \\
 & MMv1 & 0.04871 & 0.16570 & 0.07437 & 0.19704 & \text{--} & \text{--} \\
 & MMv2 & 0.06074 & 0.17914 & 0.06506 & 0.18723 & \text{--} & \text{--} \\
 & MMv3 & 0.03863 & 0.15041 & 0.05540 & 0.17284 & \text{--} & \text{--} \\
\midrule

\multirow{5}{*}{\textbf{Bearing 6}}
 & Original & 0.03739 & 0.17419 & \textbf{0.05405} & \textbf{0.21300} & \text{--} & \text{--} \\
 & MMv1 & 0.01646 & 0.11058 & 0.07797 & 0.20041 & \text{--} & \text{--} \\
 & MMv2 & \textbf{0.01199} & \textbf{0.09423} & 0.06350 & 0.18323 & \text{--} & \text{--} \\
 & MMv3 & 0.02019 & 0.12418 & 0.05437 & 0.16953 & \text{--} & \text{--} \\
\midrule

\multirow{5}{*}{\textbf{Bearing 7}}
 & Original & 0.03926 & 0.17831 & 0.04764 & 0.17043 & \text{--} & \text{--} \\
 & MMv1 & 0.03472 & 0.14534 & 0.02894 & 0.13809 & \text{--} & \text{--} \\
 & MMv2 & \textbf{0.02973} & \textbf{0.13539} & \textbf{0.02786} & \textbf{0.12547} & \text{--} & \text{--} \\
 & MMv3 & 0.03221 & 0.13822 & 0.02966 & 0.13710 & \text{--} & \text{--} \\
\bottomrule
\end{tabular}
\begin{minipage}{\textwidth}
\footnotesize
\vspace{3mm}
\textbf{Note:} Bold values indicate the minimum MSE and MAE for each bearing-condition combination.
\end{minipage}
\end{table}

\subsection{Comparison with baseline methods}
To ensure a fair comparison with existing baseline models, we employed the TF feature extraction method detailed in Algorithm~\ref{algo:tf} and trained multiple architectures, including CARLE~\cite{razzaq2025carle} and MSIDSN~\cite{zhao2023multi}, using a combined TFR from both horizontal and vertical channels. Hyperparameters for all models were fine-tuned to optimize performance. To emphasize the efficiency of our proposed architecture in terms of reduced data requirements, we trained all networks using the full 5-bearing dataset under condition 1, resulting in approximately 28\% and 48\% more training samples for the XJTU-SY and PRONOSTIA datasets, respectively. Performance metrics are summarized in Table~\ref{tab:compare_xjtu} for XJTU-SY and Table~\ref{tab:compare_pronostia} for PRONOSTIA. Our model demonstrated superior generalization capabilities, often achieving comparable or better results than baseline methods. CARLE ranked second overall, occasionally outperforming our model in select tests. 

\begin{figure}[ht!]
\centering
\includegraphics[width=0.75\textwidth]{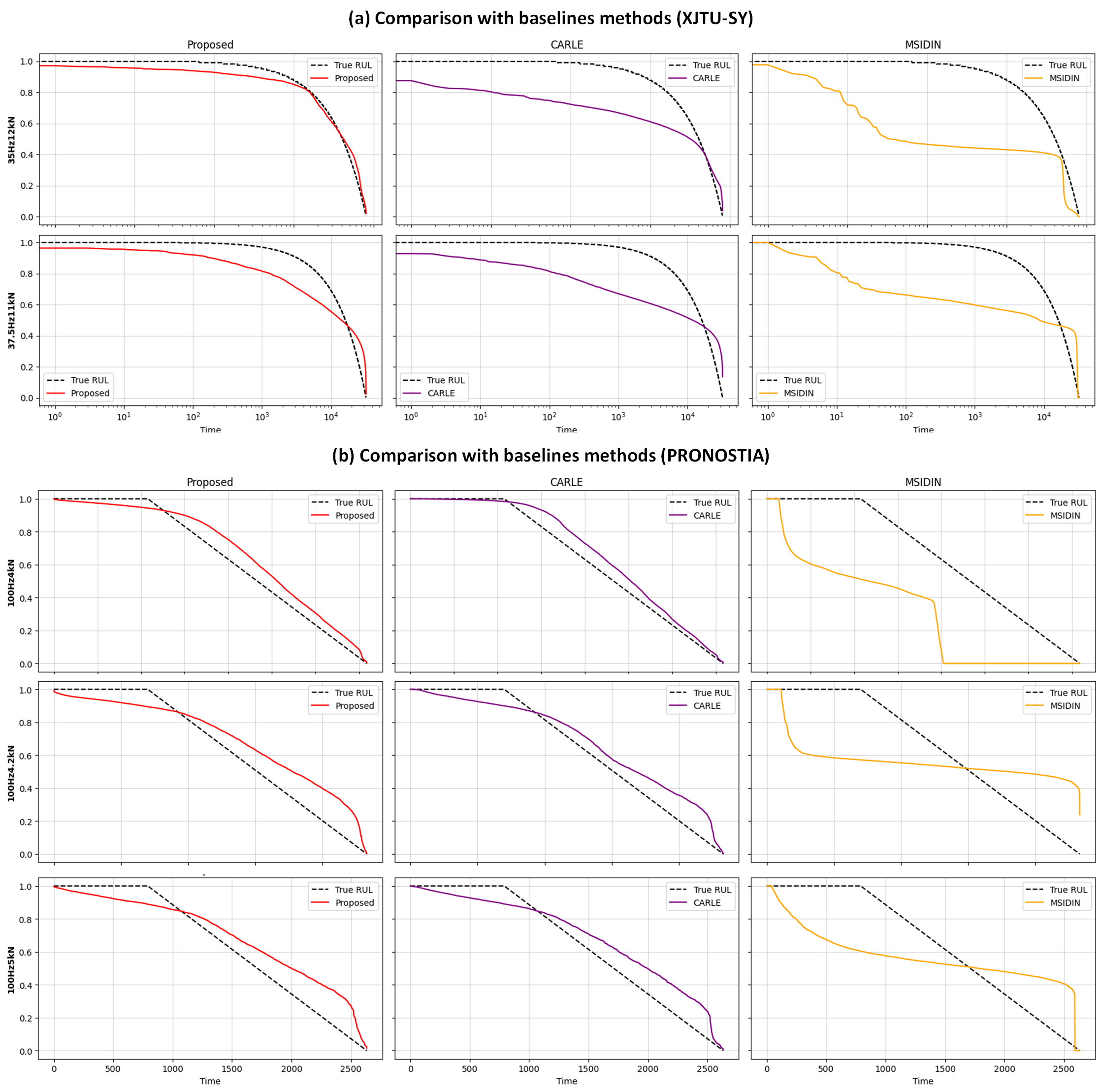}
\caption{Comparison with baseline methods across all operating conditions for \textbf{(a)} XJTU-SY \textbf{(b)} PRONOSTIA datasets. }
\label{fig:baseline}
\end{figure}

For visualization purposes, results on Bearing 1 under XJTU-SY are presented in Figure \ref{fig:baseline}(a). In condition 1, CARLE slightly outperformed our model in terms of MSE (0.01496 vs. 0.00231), but the proposed model achieved a lower MAE (0.04086 vs. 0.07871). MSIDSN ranked third with an MSE of 0.5074 and MAE of 0.17379. Under condition 2, the proposed model achieved the best performance with an MSE of 0.02006 and MAE of 0.12796, followed by MSIDSN (MSE: 0.08555, MAE: 0.24885), and CARLE (MSE: 0.09589, MAE: 0.25868).

For the PRONOSTIA dataset, results are shown in Figure \ref{fig:baseline}(b). In condition 1, our model led with an MSE of 0.00148 and MAE of 0.00328, closely followed by CARLE (MSE: 0.0015, MAE: 0.00330), while MSIDSN lagged (MSE: 2.7796, MAE: 1.2488). In condition 2, our model again ranked first (MSE: 0.01003, MAE: 0.08686), with CARLE slightly trailing (MSE: 0.0101, MAE: 0.0870), and MSIDSN third (MSE: 0.2551, MAE: 0.3937). Under condition 3, CARLE performed best (MSE: 0.0110, MAE: 0.0846), followed closely by our model (MSE: 0.01021, MAE: 0.08618), while MSIDSN remained third (MSE: 0.3339, MAE: 0.3806). These results indicate that our proposed model can achieve performance comparable to or better than existing architectures while requiring significantly less training data.

\begin{figure}[htbp]
\centering
\includegraphics[width=0.8\textwidth]{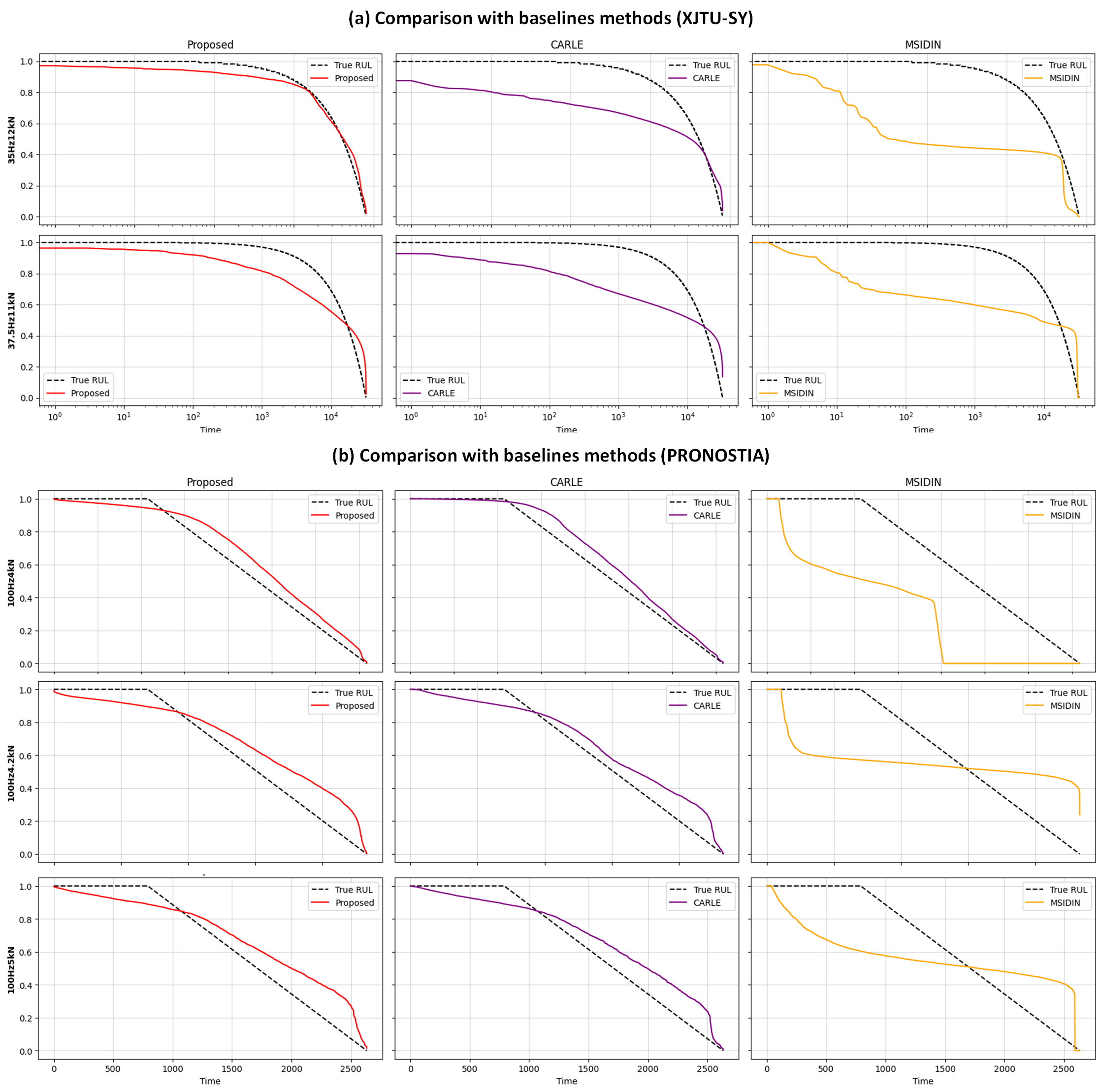}
\caption{Comparison with baseline methods for \textbf{a)} XJTU-SY; \textbf{b)} PRONOSTIA.}
\label{fig:noise}
\end{figure}

\begin{table}[htbp]
\centering
\caption{Comparison with baseline methods (XJTU-SY)}
\vspace{3mm}
\label{tab:compare_xjtu}
\footnotesize
\setlength{\tabcolsep}{5pt}
\renewcommand{\arraystretch}{0.6}
\begin{tabular}{@{}llSSSS@{}}
\toprule
\rowcolor{gray!20}
\multirow{2}{*}{\textbf{Bearing}} & \multirow{2}{*}{\textbf{Model}} &
\multicolumn{2}{c}{\textbf{35Hz12kN}} &
\multicolumn{2}{c}{\textbf{37.5Hz11kN}} \\
\cmidrule(lr){3-4} \cmidrule(lr){5-6}
 & & \textbf{MSE} & \textbf{MAE} & \textbf{MSE} & \textbf{MAE} \\
\midrule

\multirow{3}{*}{\textbf{Bearing 1}}
 & Proposed &0.00231 & \textbf{0.04086} & \textbf{0.02006} & \textbf{0.12796} \\
 & CARLE & \textbf{0.01496} & 0.07871 & 0.09589 & 0.25868 \\
 & MSIDSN & 0.05074 & 0.17379 & 0.08555 & 0.24885 \\
\midrule

\multirow{3}{*}{\textbf{Bearing 2}}
 & Proposed &  \textbf{0.00357} & \textbf{0.05182} & \textbf{0.03374} & \textbf{0.16286} \\
 & CARLE & 0.01470 & 0.06928 & 0.15661 & 0.30973 \\
 & MSIDSN & 0.05763 & 0.17848 & 0.12322 & 0.28668 \\
\midrule

\multirow{3}{*}{\textbf{Bearing 3}}
 & Proposed &  \textbf{0.00559} & \textbf{0.06174} & \textbf{0.00510} & \textbf{0.06651} \\
 & CARLE & 0.01158 & 0.06321 & 0.10200 & 0.26953 \\
 & MSIDSN & 0.04383 & 0.15704 & 0.08407 & 0.24669 \\
\midrule

\multirow{3}{*}{\textbf{Bearing 4}}
 & Proposed & \textbf{0.02337} & 0.13552 & \textbf{0.01518} & \textbf{0.11001} \\
 & CARLE & 0.02838 & \textbf{0.12417} & 0.12216 & 0.30407 \\
 & MSIDSN & 0.08448 & 0.24985 & 0.08910 & 0.25441 \\
\midrule

\multirow{3}{*}{\textbf{Bearing 5}}
 & Proposed  & \textbf{0.01039} & \textbf{0.08696} & \textbf{0.00915} & \textbf{0.07942} \\
 & CARLE & 0.01983 & 0.09124 & 0.11868 & 0.29407 \\
 & MSIDSN & 0.06368 & 0.19819 & 0.11293 & 0.28494 \\
\bottomrule
\end{tabular}
\begin{minipage}{\textwidth}
\footnotesize
\vspace{3mm}
\textbf{Note:} Bold values indicate the minimum MSE and MAE for each bearing-condition combination.
\end{minipage}
\end{table}

\begin{table}[htbp]
\centering
\caption{Comparison with baseline methods (PRONOSTIA)}
\vspace{3mm}
\label{tab:compare_pronostia}
\footnotesize
\setlength{\tabcolsep}{5pt}
\renewcommand{\arraystretch}{0.6}
\begin{tabular}{@{}llSSSSSS@{}}
\toprule
\rowcolor{gray!20}
\multirow{2}{*}{\textbf{Bearing}} & \multirow{2}{*}{\textbf{Model}} &
\multicolumn{2}{c}{\textbf{100Hz4kN}} &
\multicolumn{2}{c}{\textbf{100Hz4.2kN}} &
\multicolumn{2}{c}{\textbf{100Hz5kN}} \\
\cmidrule(lr){3-4} \cmidrule(lr){5-6} \cmidrule(lr){7-8}
 & & \textbf{MSE} & \textbf{MAE} & \textbf{MSE} & \textbf{MAE} & \textbf{MSE} & \textbf{MAE} \\
\midrule

\multirow{3}{*}{\textbf{Bearing 1}}
 & Proposed & \textbf{0.00148} & \textbf{0.03280} & \textbf{0.01003} & \textbf{0.08686} & 0.01021 & 0.08618 \\
 & CARLE & 0.0015 & 0.0330 & 0.0101 & 0.0870 & \textbf{0.0110} & \textbf{0.0846} \\ 
 & MSIDSN & 2.7796 & 1.2488 & 0.2551 & 0.3937 & 0.3339 & 0.3806 \\
 
\midrule

\multirow{3}{*}{\textbf{Bearing 2}}
 & Proposed & 0.00755 & 0.07163 & 0.00859 & 0.07791 & 0.01184 & 0.09474 \\
 & CARLE & \textbf{0.0055} & \textbf{0.0508} & \textbf{0.0054} & \textbf{0.0587} & \textbf{0.0015} & \textbf{0.0317} \\ 
 & MSIDSN & 0.2749 & 0.4065 & 0.3321 & 0.4688 & 0.3006 & 0.4427 \\
\midrule

\multirow{3}{*}{\textbf{Bearing 3}}
 & Proposed & \textbf{0.00357} & \textbf{0.05222} & \textbf{0.01174} & 0.09351 & \textbf{0.03588} & \textbf{0.17025} \\
 & CARLE & 0.0040 & 0.0530 & 0.0112 & \textbf{0.0847} & 0.1366 & 0.3230 \\ 
 & MSIDSN & 1.6971 & 0.8789 & 0.6957 & 0.5925 & 0.1596 & 0.3558 \\
\midrule

\multirow{3}{*}{\textbf{Bearing 4}}
 & Proposed & \textbf{0.00259} & \textbf{0.03638} & \textbf{0.00929} & 0.08340 & \text{--} & \text{--} \\
 & CARLE & 0.0026 & 0.0364 & 0.0124 & \textbf{0.0819} & \text{--} & \text{--} \\ 
 & MSIDSN & 0.1899 & 0.3930 & 0.3031 & 0.4149 & \text{--} & \text{--} \\
\midrule

\multirow{3}{*}{\textbf{Bearing 5}}
 & Proposed & \textbf{0.04703} & 0.18604 & \textbf{0.06076} & 0.20063 & \text{--} & \text{--} \\
 & CARLE & 0.0471 & \textbf{0.0529} & 0.0610 & \textbf{0.0807} & \text{--} & \text{--} \\ 
 & MSIDSN & 0.3049 & 0.4429 & 0.3577 & 0.4457 & \text{--} & \text{--} \\
\midrule

\multirow{3}{*}{\textbf{Bearing 6}}
 & Proposed & 0.0375 & 0.17419 & \textbf{0.05405} & 0.21300 & \text{--} & \text{--} \\
 & CARLE & \textbf{0.03739} & \textbf{0.0736} & 0.0541 & \textbf{0.0839} & \text{--} & \text{--} \\ 
 & MSIDSN & 0.5660 & 0.5515 & 0.2195 & 0.3752 & \text{--} & \text{--} \\
\midrule

\multirow{3}{*}{\textbf{Bearing 7}}
 & Proposed & 0.03926 & 0.17831 & 0.04764 & 0.17043 & \text{--} & \text{--} \\
 & CARLE & \textbf{0.0109} & \textbf{0.0778} & \textbf{0.0120} & \textbf{0.0853} & \text{--} & \text{--} \\ 
 & MSIDSN & 0.2209 & 0.3829 & 0.1657 & 0.3577 & \text{--} & \text{--} \\
\bottomrule
\end{tabular}
\begin{minipage}{\textwidth}
\footnotesize
\vspace{3mm}
\textbf{Note:} Bold values indicate the minimum MSE and MAE for each bearing-condition combination. 
\end{minipage}
\end{table}

\subsubsection{Size comparison}
\begin{figure}[ht!]
\centering
\includegraphics[width=0.95\textwidth]{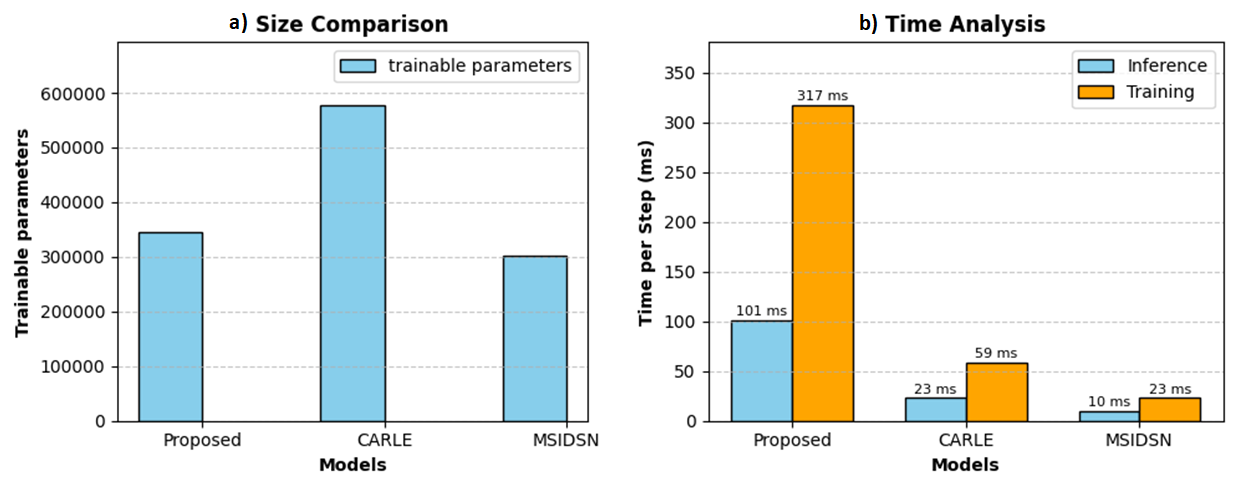}
\caption{Size comparison and Time analysis with baseline models. }
\label{fig:size}
\end{figure}

In industrial settings, the size of AI models is critical for localized implementation in PHM systems. The goal of PHM research is to reduce model size and maintain low inference times for efficient use on resource-constrained devices. The size comparison shown in Figure \ref{fig:size}(a) indicates that the proposed model falls between CARLE and MSIDSN, with trainable parameters more closely aligned with MSIDSN. However, as shown in Figure \ref{fig:size}(b), the time required to perform one training and inference step is significantly higher due to parallel computation in the Image and TF branches.

\subsection{Robustness under noisy conditions}
Evaluating the robustness and reliability of AI models under noisy conditions is essential, particularly for real-world applications where data are often affected by sensor noise, environmental variability, and system uncertainties. To simulate such conditions, we introduced controlled noise into the input data to assess the model’s stability and generalization capability beyond ideal scenarios. In our experiments, we introduced three distinct noises: \textit{Uniform Noise}, \textit{Gaussian Noise}, and \textit{salt-and-pepper noise}. Uniform noise is generated within a defined interval, specified by the lower and upper bounds (-0.003 to 0.05). For Gaussian noise, the mean ($\mu = 0$) and standard deviation (($\sigma = 0.03$)) determine the distribution’s center and spread, respectively. Salt-and-pepper noise introduces extreme values by randomly replacing data points with the maximum (salt) or minimum (pepper) values of the dataset, where probabilities of salt and pepper (10$\%$ and 20$\%$) control their occurrence rate. For visualization, Bearing 1 from each condition of both datasets was used. The impact of various noise types is illustrated in Figure~\ref{fig:noise}. In the case of the XJTU-SY dataset, shown in Figure~\ref{fig:noise}(a), the model demonstrates strong resilience to all types of noise, with outputs nearly identical to the noise-free predictions. In contrast, for the PRONOSTIA dataset, shown in Figure~\ref{fig:noise}(b), the effect of noise varies. While the model exhibits moderate robustness to salt-and-pepper noise, it performs even better under the remaining noise types in condition 1. However, in conditions 2 and 3, the presence of noise slightly degrades performance. Despite this, the model still successfully captures the long-term degradation trends, highlighting its robustness in noisy environments.

\begin{figure}[htbp]
\centering
\includegraphics[width=0.9\textwidth]{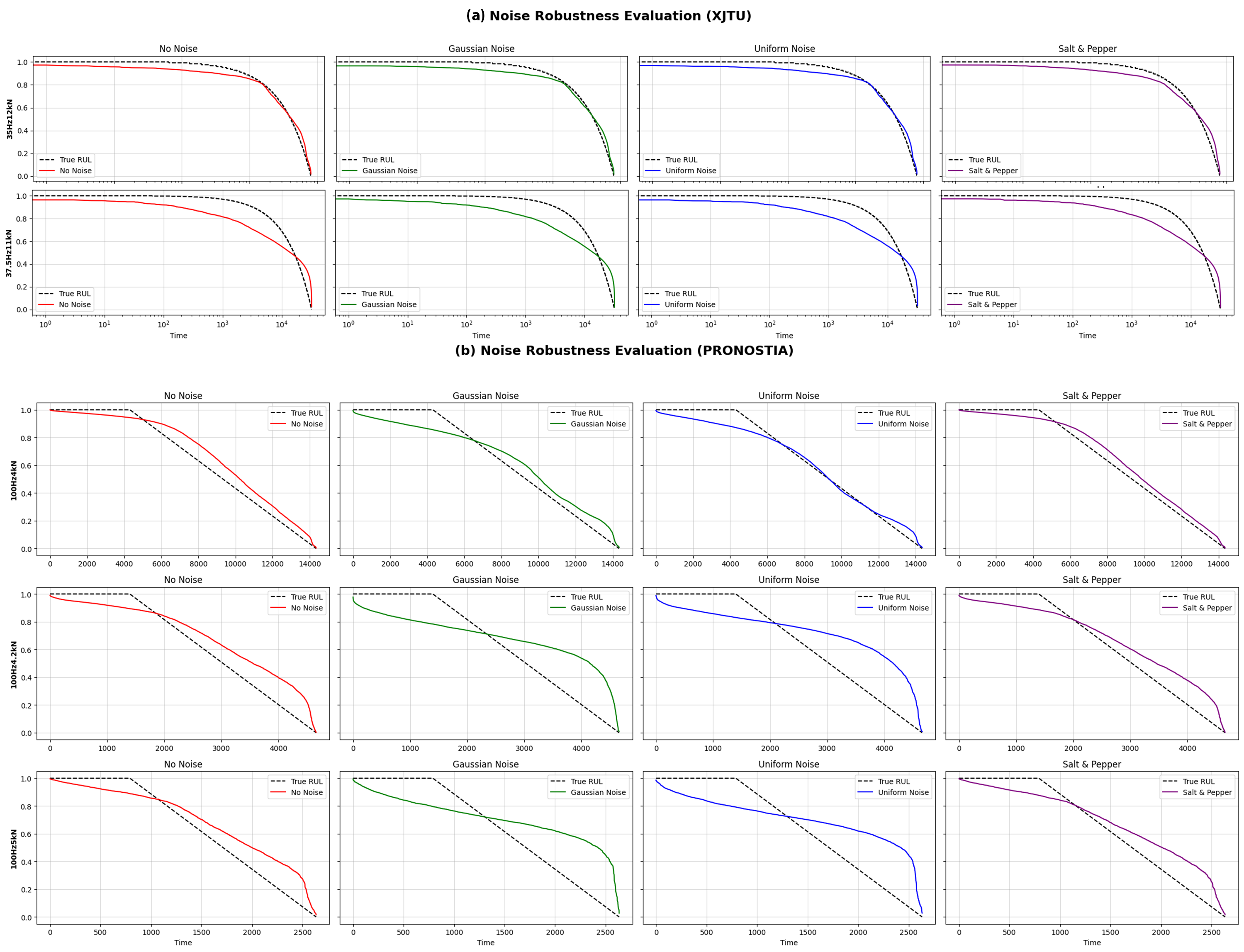}
\caption{ Noise experiment result for \textbf{a)} XJTU-SY; \textbf{b)} PRONOSTIA.}
\label{fig:noise}
\end{figure}

\subsection{LRP Explanation}
\begin{figure}[ht!]
\centering
\includegraphics[width=0.9\textwidth]{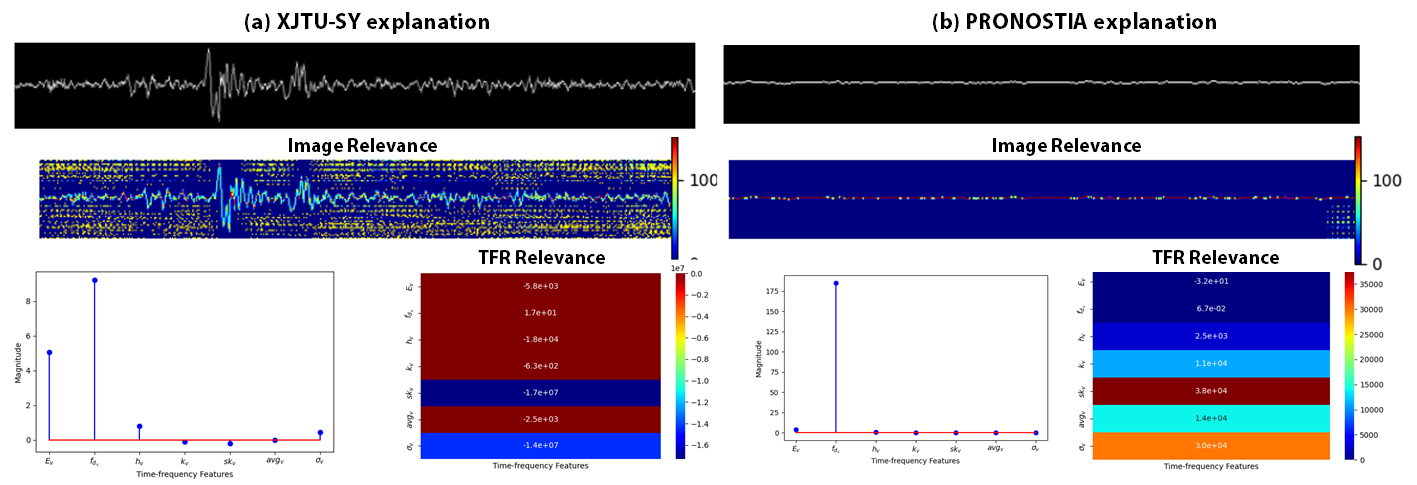}
\caption{Localized LRP explanations of multimodal AI framework with \textbf{(a)} XJTU-SY \textbf{(b)} PRONOSTIA datasets. }
\label{fig:lrp}
\end{figure}

Figure \ref{fig:lrp} presents the localized multimodal-LRP explanations for both the XJTU-SY and PRONOSTIA-trained AI. For image-branch, the LRP heatmaps suggest that the model primarily attends to complex regions of the signal, such as peaks and fluctuations, with red and yellow areas in the heatmap generally indicating higher relevance. The visualization shows that the bluish region does not strongly influence the final output, suggesting that the models selectively weigh certain signal characteristics. On the TF branch, the corresponding feature range score with overlaid relevance heatmaps. These results indicate that relevance is distributed across all extracted features, with $\sigma_v$ playing the dominant role in the prediction for both XJTU-SY and PRONOSTIA. Overall, these interpretability insights affirm that the proposed model not only performs effectively but also makes decisions based on meaningful and physically relevant features, reinforcing its trustworthiness and transparency in predictive maintenance applications.

\section{Discussion}\label{sec:discussion}

While the proposed model demonstrates strong performance, several limitations must be acknowledged to improve its practical applicability in real-world industrial scenarios. One notable challenge is the limited availability of complete run-to-failure datasets, which are often difficult to obtain in real operational environments. This scarcity can hinder the model’s ability to generalize across different machines or operating conditions. To address this, transfer learning techniques could be explored to allow the model to leverage knowledge from related domains or previously learned tasks, enhancing robustness in data-limited settings.

Another limitation lies in the relatively large size and computational demands of the current model, which may impede its deployment in resource-constrained environments such as edge devices or embedded systems. Reducing model complexity through compression methods—such as knowledge distillation, pruning, or quantization—could facilitate efficient inference while maintaining predictive performance, enabling broader deployment in real-time applications.

In terms of signal processing, the Morlet wavelet was chosen as the mother wavelet for TFR extraction due to its similarity to the bearing impulse response \cite{zhu2018estimation}. However, according to the Heisenberg-Gabor uncertainty principle, wavelet-based methods are inherently limited in their ability to achieve high resolution in both time and frequency domains simultaneously. This suggests that more advanced time-frequency techniques, such as the synchrosqueezing transform or parametric transforms, may provide more precise and informative feature representations, potentially improving diagnostic accuracy.

Furthermore, the current study does not incorporate any dedicated hyperparameter optimization strategy. Utilizing approaches such as Bayesian optimization could systematically explore the hyperparameter space and lead to better model configurations, improving both accuracy and training efficiency. Addressing these limitations through targeted research could significantly enhance the scalability, efficiency, and practical utility of the proposed framework in industrial prognostics and health management applications.

\section{Conclusion}\label{sec:conclude}
This work presents a novel multimodal-RUL framework that leverages ImR and TFR to estimate the residual operating life of rolling-element bearings. The proposed multimodal-AI network consists of three branches: (1) Image branch that captures spatial degradation patterns from ImR using dilated CNN blocks, (2) TFR branch that extracts complementary features from TFR using a similar CNN architecture as in the image branch, and (3) Fusion branch where the extracted features are concatenated and passed through an LSTM block to model temporal dependencies. Within this fusion branch, MHA mechanism selectively emphasizes the most informative features and filters out the perturbations, which are then forwarded through Linear layers for final RUL prediction. To support this architecture, we introduced a comprehensive feature engineering pipeline that transforms raw vibrational signals into ImR using the Bresenham Line Diagram and into TFR using the CWT. We also proposed a multimodal-LRP approach to provide localized explanations of the proposed AI architecture. Experimental results on the XJTU-SY and PRONOSTIA datasets show that the proposed framework not only surpasses baseline models under seen operating conditions but also generalizes well to unseen scenarios. It achieves comparable performance using 28\% and 48\% less training data on XJTU-SY and PRONOSTIA, respectively. The model also demonstrates strong robustness to input perturbations. Furthermore, multimodal-LRP analysis confirms that the model’s decision-making process is both interpretable and trustworthy, focusing on critical features across both image and TFR modalities.

\subsection{Broader Impact}
The implication of the proposed multimodal-LRP is not limited to only this use case scenario but can be utilized in any situation where multimodality plays a critical role in decision-making. This includes areas such as autonomous driving, where understanding how visual and sensor data contribute to navigation decisions is essential for safety and transparency \cite{ngiam2011multimodal}; medical diagnostics, where the combination of imaging and clinical text demands interpretable outcomes to support clinical decisions \cite{aksoy2024enhancing}; human-robot interaction, where multimodal cues (e.g., language, gestures, vision) need to be interpretable to ensure trust, collaboration, and user alignment \cite{brohan2022rt}. The ability to attribute relevance across modalities enables not only better debugging and auditing of multimodal-AI systems but also helps align model behavior with human expectations, thereby promoting ethical, transparent, and accountable deployment of AI.

\section*{Conflict of Interest}
The authors declare that they have no conflicts of interest to disclose.

\section*{Ethics Approval}
This study was conducted in accordance with ethical standards.

\section*{Funding}
The research did not receive any funding from any organization.

\section*{Data Availability}
The code for reproducibility is available at \texttt{https://github.com/itxwaleedrazzaq/PhDCode/tree/main}

\section*{Authors Contribution}
\textbf{Waleed Razzaq:} Conceptualization, Methodology, Data Curation, Writing- Original draft preparation. \textbf{Yun-Bo Zhao}: Supervision, Writing- Reviewing.

\section*{Acknowledgment}
This research was supported by the CAS-ANSO Scholarship. We acknowledge the intellectual and material contributions of the University of Science and Technology of China (USTC) and the Alliance of International Science Organizations (ANSO).

\section*{Human/Animal Participation}
No human or animal participation is involved in this research.

\bibliographystyle{unsrt}  
\bibliography{references}  

@article{ding2021remaining,
  title={Remaining useful life estimation using deep metric transfer learning for kernel regression},
  author={Ding, Yifei and Jia, Minping and Miao, Qiuhua and Huang, Peng},
  journal={Reliability Engineering \& System Safety},
  volume={212},
  pages={107583},
  year={2021},
  publisher={Elsevier}
}

@article{gabrielli2024physics,
  title={Physics-based prognostics of rolling-element bearings: The equivalent damaged volume algorithm},
  author={Gabrielli, Alberto and Battarra, Mattia and Mucchi, Emiliano and Dalpiaz, Giorgio},
  journal={Mechanical Systems and Signal Processing},
  volume={215},
  pages={111435},
  year={2024},
  publisher={Elsevier}
}

@inproceedings{gazizulin2018physics,
  title={Physics based methodology for the estimation of bearings’ remaining useful life: Physics-based models, diagnostic methods and experiments},
  author={Gazizulin, Dmitri and Klein, Renata and Bortman, Jacob},
  booktitle={Fourth European Conference of the PHM Society},
  year={2018}
}

@article{cheng2020deep,
  title={A deep learning-based remaining useful life prediction approach for bearings},
  author={Cheng, Cheng and Ma, Guijun and Zhang, Yong and Sun, Mingyang and Teng, Fei and Ding, Han and Yuan, Ye},
  journal={IEEE/ASME transactions on mechatronics},
  volume={25},
  number={3},
  pages={1243--1254},
  year={2020},
  publisher={IEEE}
}

@article{zhang2023integrated,
  title={An integrated multi-head dual sparse self-attention network for remaining useful life prediction},
  author={Zhang, Jiusi and Li, Xiang and Tian, Jilun and Luo, Hao and Yin, Shen},
  journal={Reliability Engineering \& System Safety},
  volume={233},
  pages={109096},
  year={2023},
  publisher={Elsevier}
}

@article{yang2022novel,
  title={A novel based-performance degradation indicator RUL prediction model and its application in rolling bearing},
  author={Yang, Chuangyan and Ma, Jun and Wang, Xiaodong and Li, Xiang and Li, Zhuorui and Luo, Ting},
  journal={ISA transactions},
  volume={121},
  pages={349--364},
  year={2022},
  publisher={Elsevier}
}

@article{zhu2018estimation,
  title={Estimation of bearing remaining useful life based on multiscale convolutional neural network},
  author={Zhu, Jun and Chen, Nan and Peng, Weiwen},
  journal={IEEE Transactions on Industrial Electronics},
  volume={66},
  number={4},
  pages={3208--3216},
  year={2018},
  publisher={IEEE}
}

@article{luo2022convolutional,
  title={Convolutional neural network based on attention mechanism and Bi-LSTM for bearing remaining life prediction},
  author={Luo, Jiahang and Zhang, Xu},
  journal={Applied Intelligence},
  pages={1--16},
  year={2022},
  publisher={Springer}
}

@article{patel2023non,
  title={Non-stationary neural signal to image conversion framework for image-based deep learning algorithms},
  author={Patel, Sahaj Anilbhai and Yildirim, Abidin},
  journal={Frontiers in Neuroinformatics},
  volume={17},
  pages={1081160},
  year={2023},
  publisher={Frontiers Media SA}
}

@incollection{bresenham1998algorithm,
  title={Algorithm for computer control of a digital plotter},
  author={Bresenham, Jack E},
  booktitle={Seminal graphics: pioneering efforts that shaped the field},
  pages={1--6},
  year={1998},
  publisher={NA}
}

@article{montavon2019layer,
  title={Layer-wise relevance propagation: an overview},
  author={Montavon, Gr{\'e}goire and Binder, Alexander and Lapuschkin, Sebastian and Samek, Wojciech and M{\"u}ller, Klaus-Robert},
  journal={Explainable AI: interpreting, explaining and visualizing deep learning},
  pages={193--209},
  year={2019},
  publisher={Springer}
}

@article{wang2018hybrid,
  title={A hybrid prognostics approach for estimating remaining useful life of rolling element bearings},
  author={Wang, Biao and Lei, Yaguo and Li, Naipeng and Li, Ningbo},
  journal={IEEE Transactions on Reliability},
  volume={69},
  number={1},
  pages={401--412},
  year={2018},
  publisher={IEEE}
}

@article{ngui2013wavelet,
  title={Wavelet analysis: mother wavelet selection methods},
  author={Ngui, Wai Keng and Leong, M Salman and Hee, Lim Meng and Abdelrhman, Ahmed M},
  journal={Applied mechanics and materials},
  volume={393},
  pages={953--958},
  year={2013},
  publisher={Trans Tech Publ}
}

@article{zhao2023multi,
  title={Multi-scale integrated deep self-attention network for predicting remaining useful life of aero-engine},
  author={Zhao, Ke and Jia, Zhen and Jia, Feng and Shao, Haidong},
  journal={Engineering Applications of Artificial Intelligence},
  volume={120},
  pages={105860},
  year={2023},
  publisher={Elsevier}
}

@article{lin2000feature,
  title={Feature extraction based on Morlet wavelet and its application for mechanical fault diagnosis},
  author={Lin, Jing and Qu, Liangsheng},
  journal={Journal of sound and vibration},
  volume={234},
  number={1},
  pages={135--148},
  year={2000},
  publisher={Elsevier}
}

@inproceedings{nectoux2012pronostia,
  title={PRONOSTIA: An experimental platform for bearings accelerated degradation tests.},
  author={Nectoux, Patrick and Gouriveau, Rafael and Medjaher, Kamal and Ramasso, Emmanuel and Chebel-Morello, Brigitte and Zerhouni, Noureddine and Varnier, Christophe},
  booktitle={IEEE International Conference on Prognostics and Health Management, PHM'12.},
  pages={1--8},
  year={2012},
  organization={IEEE Catalog Number: CPF12PHM-CDR}
}

@article{gao2024long,
  title={Long-term temporal attention neural network with adaptive stage division for remaining useful life prediction of rolling bearings},
  author={Gao, Pengjie and Wang, Junliang and Shi, Ziqi and Ming, Weiwei and Chen, Ming},
  journal={Reliability Engineering \& System Safety},
  volume={251},
  pages={110218},
  year={2024},
  publisher={Elsevier}
}

@article{magadan2024robust,
  title={Robust prediction of remaining useful lifetime of bearings using deep learning},
  author={Magad{\'a}n, Luis and Granda, Juan C and Su{\'a}rez, Francisco J},
  journal={Engineering Applications of Artificial Intelligence},
  volume={130},
  pages={107690},
  year={2024},
  publisher={Elsevier}
}

@article{wen2024early,
  title={Early prediction of remaining useful life for rolling bearings based on envelope spectral indicator and Bayesian filter},
  author={Wen, Haobin and Zhang, Long and Sinha, Jyoti K},
  journal={Applied Sciences},
  volume={14},
  number={1},
  pages={436},
  year={2024},
  publisher={MDPI}
}

@article{bach2015pixel,
  title={On pixel-wise explanations for non-linear classifier decisions by layer-wise relevance propagation},
  author={Bach, Sebastian and Binder, Alexander and Montavon, Gr{\'e}goire and Klauschen, Frederick and M{\"u}ller, Klaus-Robert and Samek, Wojciech},
  journal={PloS one},
  volume={10},
  number={7},
  pages={e0130140},
  year={2015},
  publisher={Public Library of Science San Francisco, CA USA}
}

@article{samek2016interpreting,
  title={Interpreting the predictions of complex ML models by layer-wise relevance propagation},
  author={Samek, Wojciech and Montavon, Gr{\'e}goire and Binder, Alexander and Lapuschkin, Sebastian and M{\"u}ller, Klaus-Robert},
  journal={arXiv preprint arXiv:1611.08191},
  year={2016}
}

@inproceedings{ngiam2011multimodal,
  title={Multimodal deep learning.},
  author={Ngiam, Jiquan and Khosla, Aditya and Kim, Mingyu and Nam, Juhan and Lee, Honglak and Ng, Andrew Y and others},
  booktitle={ICML},
  volume={11},
  pages={689--696},
  year={2011}
}

@inproceedings{aksoy2024enhancing,
  title={Enhancing Image-to-Text Generation in Radiology Reports through Cross-modal Multi-Task Learning},
  author={Aksoy, Nurbanu and Ravikumar, Nishant and Sharoff, Serge},
  booktitle={Proceedings of the 2024 Joint International Conference on Computational Linguistics, Language Resources and Evaluation (LREC-COLING 2024)},
  pages={5977--5985},
  year={2024}
}

@article{brohan2022rt,
  title={Rt-1: Robotics transformer for real-world control at scale},
  author={Brohan, Anthony and Brown, Noah and Carbajal, Justice and Chebotar, Yevgen and Dabis, Joseph and Finn, Chelsea and Gopalakrishnan, Keerthana and Hausman, Karol and Herzog, Alex and Hsu, Jasmine and others},
  journal={arXiv preprint arXiv:2212.06817},
  year={2022}
}

@inproceedings{ni2024dynamic,
  title={Dynamic Simulation and Vibration Analysis of Rolling Bearing with Outer Ring Defects},
  author={Ni, Wenjun and Zhang, Chang},
  booktitle={International Conference on Mechanical Design and Simulation},
  pages={705--713},
  year={2024},
  organization={Springer}
}

@article{guo2023dynamics,
  title={Dynamics modeling and analysis of rolling bearings variable stiffness system with local faults},
  author={Guo, Baoliang and Wu, Wenlong and Zheng, Jianxiao and He, Yumin and Zhang, Jinhua},
  journal={Machines},
  volume={11},
  number={6},
  pages={609},
  year={2023},
  publisher={MDPI}
}

@article{razzaq2025carle,
  title={CARLE: a hybrid deep-shallow learning framework for robust and explainable RUL estimation of rolling element bearings},
  author={Razzaq, Waleed and Zhao, Yun-Bo},
  journal={Soft Computing},
  volume={29},
  number={23},
  pages={6269--6292},
  year={2025},
  publisher={Springer}
}

\end{document}